\documentclass{tlp}
\usepackage{aopmath}

\newtheorem{definition}{Definition} 
\newtheorem{example}{Example} 

\usepackage{amssymb}
\usepackage{graphicx}
\usepackage{alltt}
\usepackage{xspace}
\usepackage{url}
\usepackage{subfigure}
\usepackage{color}

\newcommand{\domc}{{\bf C}}
\newcommand{\dep}{\Sigma}
\newcommand{\variables}{{\bf V}}
\newcommand{\ins}[1]{\mathbf{#1}}
\newcommand{\insX}{\ins{X}}
\newcommand{\insY}{\ins{Y}}
\newcommand{\insZ}{\ins{Z}}
\newcommand{\insW}{\ins{W}}
\newcommand{\tup}{\ins{t}}
\newcommand{\R}{\mathcal{R}}
\newcommand{\Sch}{\Sigma}
\newcommand{\atom}[1]{\underline{#1}}

\newcommand{\var}[1]{\mathit{var}(#1)}

\newcommand{\qed}{\hfill \ensuremath{\Box}}
\newcommand{\nop}[1]{}

\newcommand{\pruning}{\textit{Pruning}\xspace}
\newcommand{\bertossi}{\textit{BB}\xspace}
\newcommand{\tplp}{\textit{MRT}\xspace}

\newcommand{\wasptwo}{WASP 2.0\xspace}
\newcommand{\wasp}{WASP\xspace}
\newcommand{\clasp}{clasp\xspace}

\newcommand{\myParagraph}[1]{\medskip \noindent\textit{#1}}

\newcommand{\spaceBeforeSec}{\vspace{-3mm}}
\newcommand{\spaceAfterSec}{\vspace{0mm}}

\newcommand{\spaceBeforeSubSec}{\vspace{-3mm}}
\newcommand{\spaceAfterSubSec}{\vspace{-1mm}}

\newcommand{\spaceBeforeEnv}{\vspace{-1.5mm}}
\newcommand{\spaceAfterEnv}{\vspace{-1.5mm}}

\begin{document}
\bibliographystyle{acmtrans}


\title[Taming Primary Key Violations to Query Large Inconsistent Data]{Taming Primary Key Violations to Query \\ Large Inconsistent Data via ASP}

\author[M. Manna and F. Ricca and G. Terracina]
{MARCO MANNA and FRANCESCO RICCA and GIORGIO TERRACINA\\
Department of Mathematics and Computer Science\\
University of Calabria, Italy\\
E-mail: \{manna,ricca,terracina\}@mat.unical.it}

\maketitle

\begin{abstract}
Consistent query answering over a database that violates primary key constraints
is a classical hard problem in database research that has been traditionally dealt with logic programming.
However, the applicability of existing logic-based solutions is restricted to data sets of moderate size.
This paper presents a novel decomposition and pruning strategy that reduces, in polynomial time,
the problem of computing the consistent answer to a conjunctive query over a database subject to primary key constraints to a collection of smaller problems of the same sort that can be solved independently.
The new strategy is naturally modeled and implemented using Answer Set Programming (ASP).
An experiment run on benchmarks from the database world prove
the effectiveness and efficiency of our ASP-based approach also on large data sets.
To appear in Theory and Practice of Logic Programming (TPLP), Proceedings of ICLP 2015.
\end{abstract}

\begin{keywords}
Inconsistent Databases,
Primary Key Constraints,
Consistent Query Answering,
ASP
\end{keywords}

\section{Introduction}\label{sec:introduction}

Integrity constraints provide means for ensuring that database evolution does not result in a loss of consistency or in a discrepancy with the intended model of the application domain~\cite{AbHV95}.
A relational database that do not satisfy some of these constraints is said to be inconsistent.
In practice it is not unusual that one has to deal with inconsistent data~\cite{BeHS05}, and when a conjunctive query (CQ) is posed to an inconsistent database, a natural problem arises that can be formulated as: \emph{How to deal with inconsistencies to answer the input query in a consistent way?}
This is a classical problem in database research and different approaches have been proposed in the literature.
One possibility is to clean the database~\cite{ElIV07} and work on one of the possible coherent states;
another possibility is to be tolerant of inconsistencies by leaving intact the database
and computing answers that are ``consistent with the integrity constraints''~\cite{ArBC99,Bert11}.

In this paper, we adopt the second approach -- which has been proposed by \citeNS{ArBC99}
under the name of \emph{consistent query answering} (CQA) -- and focus on the relevant class of \emph{primary key} constraints.
Formally, in our setting: 
$(1)$ a database $D$ is \emph{inconsistent} if there are at least two tuples of the same relation that agree on their primary key;
$(2)$ a \emph{repair} of $D$ is any maximal consistent subset of $D$; and
$(3)$ a tuple $\tup$ of constants is in the \emph{consistent answer} to a CQ $q$ over $D$
if and only if, for each repair $R$ of $D$, tuple $\tup$ is in the (classical) answer to $q$ over $R$.
Intuitively, the original database is (virtually) repaired by applying a minimal number of corrections (deletion of tuples with the same primary key), while the consistent answer collects the tuples that can be retrieved in every repaired instance.

CQA under primary keys is coNP-complete in data complexity \cite{ArBC03}, when both the relational schema and the query are considered fixed.
%
Due to its complex nature, traditional RDBMs 
are inadequate to solve the problem alone via SQL without focusing on restricted classes of CQs \cite{ArBC99,FuFM05,FuMi07,Wijs09,Wijs12}.
Actually, in the unrestricted case, CQA has been traditionally dealt with logic programming~\cite{GrGZ01,ArBC03,BaBe03,EFGL03,GrGZ03,MaRT13}.
However, it has been argued~\cite{KoPT13} that the practical applicability of logic-based approaches is restricted to data sets of moderate size. Only recently, an approach based on Binary Integer Programming~\cite{KoPT13} has revealed good performances on large databases (featuring up to one million tuples per relation) with primary key violations.

In this paper, we demonstrate that logic programming can still be effectively used for computing consistent answers over large relational databases. 
We design a novel decomposition strategy that reduces (in polynomial time)
the computation of the consistent answer to a CQ over a database subject to primary key constraints
into a collection of smaller problems of the same sort.
At the core of the strategy is a cascade pruning mechanism that dramatically reduces the number of key violations that have to be handled to answer the query.

Moreover, we implement the new strategy using Answer Set Programming (ASP) \cite{GeLi91,BrET11}, and we prove empirically the effectiveness of our ASP-based approach on existing benchmarks from the database world.
In particular, we compare our approach with some classical \cite{BaBe03} and optimized \cite{MaRT13} encodings of CQA in ASP that were presented in the literature.
%
The experiment empirically demonstrate that our logic-based approach implements CQA efficiently on large data sets, and can even perform better than state-of-the-art methods.

\spaceBeforeSec
\section{Preliminaries}\label{sec:preliminaries}
\spaceAfterSec
We are given two disjoint countably infinite sets
of \emph{terms} denoted by $\domc$ and $\variables$
and called \emph{constants} and \emph{variables}, respectively.
We denote
by $\insX$ sequences (or sets, with a slight abuse of notation) of variables $X_1,\ldots,X_n$, and
by ${\bf t}$ sequences of terms $t_1,\ldots,t_n$.
We also denote by $[n]$ the set $\{1,\ldots,n\}$, for any $n \geqslant 1$.
Given a sequence ${\bf t} = t_1,\ldots,t_n$ of terms and a set $S = \{p_1,\ldots,p_k\} \subseteq [n]$,
${\bf t}|_{S}$ is the subsequence $t_{p_1},\ldots,t_{p_k}$.
For example, if ${\bf t} = t_1,t_2,t_3$ and $S = \{1,3\}$,
then ${\bf t}|_{S} = t_1,t_3$.

A (\emph{relational}) \emph{schema} is a triple $\langle \R, \alpha, \kappa \rangle$
where $\R$ is a finite set of \emph{relation symbols} (or \emph{predicates}),
$\alpha : \R \rightarrow \mathbb{N}$ is a function associating an \emph{arity} to each predicate,
and $\kappa : \R \rightarrow 2^\mathbb{N}$ is a function that associates, to each $r \in \R$,
a nonempty set of positions from $[\alpha(r)]$, which represents the \emph{primary key} of $r$.
Moreover, for each relation symbol $r \in \R$ and for each position $i \in [\alpha(r)]$, $r[i]$
denotes the $i$-th \emph{attribute} of $r$.
Throughout, let $\Sch = \langle \R, \alpha, \kappa \rangle$ denote a
relational schema.
An \emph{atom} (over $\Sch$) is an expression of the form $r(t_1,\ldots,t_n)$,
where $r \in \R$, and $n = \alpha(r)$.
An atom is called a \emph{fact} if all of its terms are constants of $\domc$.
Conjunctions of atoms are often identified with the sets of their atoms.
For a set $A$ of atoms, the variables occurring in
$A$ are denoted by $\var{A}$.
A \emph{database} $D$ (over $\Sch$) is a finite set of facts over $\Sigma$.
Given an atom $r({\bf t}) \in D$, we denote by $\hat{{\bf t}}$ the sequence ${\bf t}|_{\mathit{\kappa}(r)}$.
%
We say that $D$ is \emph{inconsistent} (w.r.t. $\Sch$) if it contains
two different atoms of the form $r({\bf t}_1)$ and $r({\bf t}_2)$ such that
$\hat{{\bf t}}_1 = \hat{{\bf t}}_2$. Otherwise, it is \emph{consistent}.
A \emph{repair} $R$ of $D$ (w.r.t. $\Sch$) is any maximal consistent subset of $D$.
The set of all the repairs of $D$ is denoted by $\mathit{rep}(D,\Sch)$.

A \emph{substitution} is a mapping $\mu : \domc \cup \variables \rightarrow \domc \cup \variables$ which is
the identity on $\domc$.
%
Given a set $A$ of atoms, \mbox{$\mu(A) = \{r(\mu(t_1),\ldots,\mu(t_n))~:~r(t_1,\ldots,t_n) \in A\}$}.
The restriction of $\mu$ to a set $S \subseteq \domc \cup \variables$, is denoted by $\mu|_{S}$.
%
A \emph{conjunctive query} (CQ) $q$ (over $\Sch$)
is an expression of the form $\exists \insY \, \varphi(\insX,\insY)$, where
$\insX \cup \insY$ are variables of $\variables$, and
$\varphi$ is a conjunction of atoms (possibly with constants) over $\Sch$.
To highlight the free variables of $q$, we often write $q(\insX)$ instead of $q$.
If $\insX$ is empty, then $q$ is called a \emph{Boolean conjunctive query} (BCQ).
Assuming that $\insX$ is the sequence $X_1,\ldots,X_n$,
the \emph{answer} to $q$ over a database $D$, denoted $q(D)$, is the set of all
$n$-tuples $\langle t_1,\ldots,t_n \rangle \in \domc^n$ for which there exists a
substitution $\mu$ such that $\mu(\varphi(\insX,\insY)) \subseteq D$ and
$\mu(X_i) = t_i$, for each $i \in [n]$. A BCQ is \emph{true} in $D$,
denoted $D \models q$, if $\langle \rangle \in q(D)$.
The \emph{consistent answer} to a CQ $q(\insX)$ over a database $D$
(w.r.t. $\Sch$), denoted $\mathit{ans}(q,D,\Sch)$, is the set of
tuples $\bigcap_{R \in \mathit{rep}(D,\Sch)} q(R)$.
Clearly, $\mathit{ans}(q,D,\Sch) \subseteq q(D)$ holds.
A BCQ $q$ is \emph{consistently true} in a database $D$ (w.r.t. $\Sch$),
denoted $D \models_{\Sch} q$, if $\langle \rangle  \in \mathit{ans}(q,D,\Sch)$.
%

%


\spaceBeforeSec
\section{Dealing with Large Datasets}\label{sec:pruning}
\spaceAfterSec

To deal with large inconsistent data, we design a strategy that reduces in polynomial time the problem of computing the consistent answer to a CQ over a database subject to primary key constraints
to a collection of smaller problems of the same sort.
To this end, we exploit the fact that the former problem
is logspace Turing reducible to the one of deciding whether a BCQ
is consistently true (recall that the consistent answer to a CQ is a subset of its answer). 
%
%
Hence, given a database $D$ over a schema $\Sch$,
and a BCQ $q$, we would like to identify a set $F_1,\ldots,F_k$ of pairwise disjoint subsets of $D$, called fragments, such that: {\em $D \models_\Sch q$ iff there is $i \in [k]$ such that $F_i \models_\Sch q$}.
At the core of our strategy we have:
$(1)$ a cascade pruning mechanism to reduce the number of ``crucial'' inconsistencies, and
$(2)$ a technique to identify a suitable set of fragments from any (possibly unpruned) database.
For the sake of presentation, we start with principle $(2)$.
In the last two subsections, we provide
complementary techniques to further reduce
the number of inconsistencies to be handled for answering the original CQ.
The proofs of this section are given in~\ref{sec:app-proofs}.


\spaceBeforeSubSec
\subsection{Fragments Identification}\label{sec:fragments}
\spaceAfterSubSec
Given a database $D$, a \emph{key component} $K$ of $D$ is any maximal subset of $D$ such that
if $r_1({\bf t}_1)$ and $r_2({\bf t}_2)$ are in $K$, then both $r_1 = r_2$ and $\hat{{\bf t}}_1 = \hat{{\bf t}}_2$ hold.
Namely, $K$ collects only atoms that agree on their primary key.
Hence, the set of all key components of $D$, denoted by $\mathit{comp}(D,\Sch)$, forms a partition of $D$.
If a key component is a singleton, then it is called \emph{safe}; otherwise it is \emph{conflicting}.
Let $\mathit{comp}(D,\Sch) = \{K_1,\ldots,K_n\}$. It can be verified that
$\mathit{rep}(D,\Sch) = \{\{\atom{a}_1,\ldots,\atom{a}_n\}~:~\atom{a}_1 \in K_1,\ldots,\atom{a}_n \in K_n\}$.
Let us now fix throughout this section a BCQ $q$ over $\Sch$.
For a repair $R \in\mathit{rep}(D,\Sch)$, if $q$ is true in $R$, then there is a
substitution $\mu$ such that $\mu(q) \subseteq R$. But since $R \subseteq D$,
it also holds that $\mu(q) \subseteq D$.
Hence, \mbox{$\mathit{sub}(q,D) = \{\mu|_{\var{q}}~:~\mu$ is a substitution and $\mu(q)\subseteq D\}$} is an overestimation of the substitutions that map $q$ to the repairs of $D$.

\begin{figure}[t!] \centering
\includegraphics[scale=1]{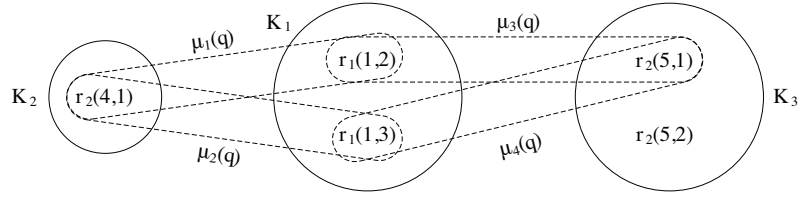}
\caption{Conflict-join hypergraph.}
\label{fig:repairs}
\end{figure}

Inspired by the notions of conflict-hypergraph \cite{ChMa05} and conflict-join graph \cite{KoPe12},
we now introduce the notion of conflict-join hypergraph.
Given a database $D$, the \textit{conflict-join hypergraph} of $D$ (w.r.t. $q$ and $\dep$)
is denoted by $H_D = \langle D, E \rangle$, where $D$ are the vertices, and $E$ are the hyperedges partitioned in
$E_q = \{\mu(q)~:~\mu \in \mathit{sub}(q,D)\}$ and $E_\kappa = \{K~:~K \in \mathit{comp}(D,\Sch)\}$.
A \emph{bunch} $B$ of vertices of $H_D$ is any minimal nonempty subset of $D$ such that, for each $e \in E$,
either $e \subseteq B$ or $e \cap B = \emptyset$ holds. 
%
Intuitively, every edge of $H_D$ collects the atoms in a key component of $D$
or the atoms in $\mu(q)$, for some $\mu \in \mathit{sub}(q,D)$.
Moreover, each bunch collects the vertices of some connected component of $H_D$.
Before we proceed further, let us fix these preliminary notions
with the aid of the following example.

\spaceBeforeEnv
\begin{example}\label{ex:hypergraph}
Consider the schema $\Sch = \langle \R, \alpha, \kappa \rangle$,
where $\R = \{r_1,r_2\}$, $\alpha(r_1) = \alpha(r_2) = 2$, and
$\kappa(r_1) = \kappa(r_2) = \{1\}$.
Consider also the database $D = \{r_1(1,2),$ $r_1(1,3),$ $r_2(4,1),$ $r_2(5,1)$, $r_2(5,2)\}$,
and the BCQ $q = r_1(X,Y), r_2(Z,X)$.
The conflicting components of $D$ are
$K_1 = \{r_1(1,2),$ $r_1(1,3)\}$ and
$K_3 = \{r_2(5,1)$, $r_2(5,2)\}$, while
its safe component is $K_2 = \{r_2(4,1)\}$.
The repairs of $D$ are
$R_1 = \{r_1(1,2),$ $r_2(4,1),$ $r_2(5,1)\}$,
$R_2 = \{r_1(1,2),$ $r_2(4,1),$ $r_2(5,2)\}$,
$R_3 = \{r_1(1,3),$ $r_2(4,1),$ $r_2(5,1)\}$, and
$R_4 = \{r_1(1,3),$ $r_2(4,1),$ $r_2(5,2)\}$.
Moreover, $\mathit{sub}(q,D)$ contains the substitutions:
$\mu_1 = \{X \mapsto 1, Y \mapsto 2, Z \mapsto 4\}$,
$\mu_2 = \{X \mapsto 1, Y \mapsto 3, Z \mapsto 4\}$,
$\mu_3 = \{X \mapsto 1, Y \mapsto 2, Z \mapsto 5\}$, and
$\mu_4 = \{X \mapsto 1, Y \mapsto 3, Z \mapsto 5\}$.
The conflict-join hypergraph $H_D = \langle D, E \rangle$ is depicted in Figure \ref{fig:repairs}.
Solid (resp., dashed) edges form the set $E_\kappa$ (resp., $E_q$).
Since $\mu_1$ maps $q$ to $R_1$ and $R_2$,
and $\mu_2$ maps $q$ to $R_3$ and $R_4$,
we conclude that $D \models_\Sch q$.
Finally, $D$ is the only bunch of $H_D$. \qed
\end{example}
\spaceAfterEnv

%
%

In Example \ref{ex:hypergraph} we observe that $K_3$ can be safely ignored in the evaluation of $q$.
In fact, even if both $\mu_3(q)$ and $\mu_4(q)$
contain an atom of $K_3$, $\mu_1$ and $\mu_2$
are sufficient to prove that $q$ is consistently true.
This might suggest to focus only on the set $F = K_1 \cup K_2$, and on its
repairs $\{r_1(1,2),$ $r_2(4,1)\}$ and $\{r_1(1,3),$ $r_2(4,1)\}$.
Also, since $F \models_\Sch q$, $F$ represents the ``small'' fragment
of $D$ that we need to evaluate $q$.
The practical advantage of considering $F$ instead of $D$ should be already clear:
$(1)$ the repairs of $F$ are smaller than the repairs of $D$; and
$(2)$ $F$ has less repairs than $D$.
%
%
%
We are now ready to introduce the the formal notion of fragment.

\spaceBeforeEnv
\begin{definition}\label{def:fragment}
Consider a database $D$.
For any 
set $C \subseteq \mathit{comp}(D,\Sch)$
of key components of $D$, we say that
the set $F = \bigcup_{K \in C} K$ is a (\emph{well-defined}) \emph{fragment} of $D$. \qed
\end{definition}
\spaceAfterEnv

According to Definition \ref{def:fragment},
the set $F = K_1 \cup K_2$ in Example \ref{ex:hypergraph}
is a fragment of $D$.
The following proposition, states a useful property
that holds for any fragment.

\spaceBeforeEnv
\begin{proposition}\label{prop:fragment}
Consider a database $D$, and two fragments $F_1 \subseteq F_2$ of $D$.
If $F_1 \models_\Sch q$, then $F_2 \models_\Sch q$.
\end{proposition}
\spaceAfterEnv

By Definition \ref{def:fragment}, $D$ is indeed a fragment of itself.
Hence, if $q$ is consistently true,
then there is always the fragment $F = D$ such that $F \models_\Sch q$.
But now the question is: \emph{How can we identify a convenient set of fragments of $D$}?
The naive way would be to use as fragments the bunches of $H_D$.
%
%
Soundness is guaranteed by Proposition \ref{prop:fragment}.
%
Regarding completeness, we rely on the following result.

\spaceBeforeEnv
\begin{theorem}\label{thm:bunches}
Consider a database $D$. If $D \models_\Sch q$, then
there is a bunch $B$ of $H_D$ s.t. $B \models_\Sch q$.
\end{theorem}
\spaceAfterEnv

By combining Proposition \ref{prop:fragment} with Theorem \ref{thm:bunches}
we are able to reduce, in polynomial time, the original problem into a collection of smaller ones of the same sort.
%

\spaceBeforeSubSec
\subsection{The Cascade Pruning Mechanism}\label{sec:pruningcascade}
\spaceAfterSubSec
The technique proposed in the previous section alone
is not sufficient to deal with large data sets.
In fact, since it considers all the bunches of the conflict-join hypergraph,
it unavoidably involves the entire database.
%
%
To strengthen its effectiveness,
we need an algorithm that realizes, for instance,
that $K_3$ is ``redundant'' in Example \ref{ex:hypergraph}.
%
%
%
But before that, let us define formally what we mean by the term redundant.


\spaceBeforeEnv
\begin{definition}\label{def:redundant}
A key component $K$ of a database $D$ is called \emph{redundant}
(w.r.t. $q$) if the following condition is satisfied:
for each fragment $F$ of $D$, $F \models_\Sch q$ implies $F\setminus K \models_\Sch q$. \qed
%
\end{definition}
\spaceAfterEnv
The above definition states that
a key component is redundant independently from the fact that some other
key component is redundant or not.
Therefore:

\spaceBeforeEnv
\begin{proposition}\label{prop:redundancyClosure}
Consider a database $D$ and a set $C$ of redundant components of $D$.
It holds that $D \models_\Sch q$ iff $\left(D \setminus \bigcup_{K \in C} K\right)  \models_\Sch q$.
\end{proposition}
\spaceAfterEnv

In light of Proposition~\ref{prop:redundancyClosure},
if we can identify all the redundant components of $D$, then after removing from $D$ all these components, what remains is either:
$(1)$ a nonempty set of (minimal) bunches, each of which entails consistently $q$ whenever $D \models_\Sch q$; or
$(2)$ the empty set, whenever $D \not\models_\Sch q$. More formally:

\spaceBeforeEnv
\begin{proposition}\label{prop:TrueQueryHasNonRedundantComponent}
Given a database $D$, each key component of $D$ is redundant iff $D \not\models_\Sch q$.
\end{proposition}
\spaceAfterEnv

%

However, assuming that $\textsc{ptime} \neq \textsc{np}$,
any algorithm for the identification of all the redundant components of $D$
cannot be polynomial because, otherwise,
we would have a polynomial procedure for solving the original problem.
Our goal is therefore to identify sufficient conditions
to design a pruning mechanism that detects in polynomial time
as many redundant conflicting components as possible.
To give an intuition of our pruning mechanism,
we look again at Example \ref{ex:hypergraph}.
Actually, $K_3$ is redundant because it contains an atom, namely $r_2(5,2)$,
that is not involved in any substitution (see Figure \ref{fig:repairs}).
Assume now that this is the criterion that we use to identify redundant components.
Since, by Definition \ref{def:redundant}, we know that $D \models_\Sch q$ iff $D \setminus K_3 \models_\Sch q$,
this means that we can now forget about $D$ and consider only $D' = K_1 \cup K_2$.
But once we focus on $\mathit{sub}(q,D')$, we realize that it contains only $\mu_1$ and $\mu_2$.
Then, a smaller number of substitutions in $\mathit{sub}(q,D')$ w.r.t. those in $\mathit{sub}(q,D)$ motivates us to reapply our criterion. Indeed, there could also be some atom in $D'$ not involved in any of the substitutions of $\mathit{sub}(q,D')$.
This is not the case in our example since the atoms in $D'$ are covered by
$\mu_1(q)$ or $\mu_2(q)$.
However, in general, in one or more steps, we can identify more and more
redundant components.
We can now state the main result of this section.

\spaceBeforeEnv
\begin{theorem}\label{thm:pruning}
Consider a database $D$, and a key component $K$ of $D$.
Let $H_D = \langle D, E \rangle$ be the conflict-join hypergraph of $D$.
If $K \setminus \bigcup_{e \in E_q} e \neq \emptyset$,
then $K$ is redundant.
\end{theorem}
\spaceAfterEnv

In what follows, a redundant component that can be identified
via Theorem~\ref{thm:pruning} is called
\emph{strongly redundant}.
As discussed just before Theorem~\ref{thm:pruning}, an indirect effect of removing a redundant component $K$ from $D$ is that all the substitutions in the set
\mbox{$S =\{\mu \in \mathit{sub}(q,D)~:~\mu(q) \cap K \neq \emptyset\}$}
can be in a sense ignored. In fact, $\mathit{sub}(q,D \setminus K) = \mathit{sub}(q,D) \setminus S$.
Whenever a substitution can be safely ignored,
we say that it is unfounded.
Let us formalize this new notion in the following definition.


\spaceBeforeEnv
\begin{definition}\label{def:unfoundedSub}
Consider a database $D$.
A substitution $\mu$ of $\mathit{sub}(q,D)$ is \emph{unfounded}
if: 
for each fragment $F$ of $D$, $F \models_\Sch q$ implies that,
for each repair $R \in \mathit{rep}(F,\Sch)$,
there exists a substitution $\mu' \in \mathit{sub}(q,R)$ different from $\mu$
such that $\mu'(q) \subseteq R$. \qed
\end{definition}
\spaceAfterEnv
We now show how to detect as many unfounded substitutions as possible.

\spaceBeforeEnv
\begin{theorem}\label{thm:suffUnfoundedSub}
Consider a database $D$, and a substitution $\mu \in \mathit{sub}(q,D)$.
If there exists a redundant component $K$ of $D$
such that $\mu(q) \cap K \neq \emptyset$, then
$\mu$ is unfounded.
\end{theorem}
\spaceAfterEnv

Clearly, Theorem~\ref{thm:suffUnfoundedSub} alone is not helpful since
it relies on the identification of redundant components. However,
if combined with Theorem~\ref{thm:pruning},
it forms the desired cascade pruning mechanism.
%
%
To this end, we call \emph{strongly unfounded} an unfounded substitution
that can be identified by applying Theorem~\ref{thm:suffUnfoundedSub} by only considering
strongly redundant components.
Hereafter, let us denote by $\mathit{sus}(q,D)$
the subset of $\mathit{sub}(q,D)$ containing only strongly unfounded substitutions.
Hence, both substitutions $\mu_3$ and $\mu_4$ in Example \ref{ex:hypergraph}
are strongly unfounded, since $K_3$ is strongly redundant.
Moreover, we reformulate the statement of Theorem \ref{thm:pruning}
by exploiting the notion of strongly unfounded substitution,
and the fact that the set $K \setminus \bigcup_{e \in E_q} e$ is nonempty
if and only if there exists an atom $\atom{a} \in K$ such that
the set $\{\mu \in \mathit{sub}(q,D)~:~\atom{a} \in \mu(q)\}$ --
or equivalently the set $\{e \in E_q~:~\atom{a} \in e\}$ --
is empty.
For example, according to Figure~\ref{fig:repairs},
the set $K_3 \setminus \bigcup_{e \in E_q} e$ is nonempty since
it contains the atom $r_2(5,2)$. But this atoms makes the set
$\{\mu \in \mathit{sub}(q,D)~:~r_2(5,2) \in \mu(q)\}$ empty since
no substitution of $\mathit{sub}(q,D)$ (or no hyperedge of $E_q$)
involves $r_2(5,2)$.


\spaceBeforeEnv
\begin{proposition}\label{prop:stronglyRedundant}
A key component $K$ of $D$ is strongly redundant if
there is an atom $\atom{a} \in K$ such that
one of the two following conditions is satisfied:
(1) $\{\mu \in \mathit{sub}(q,D)~:~\atom{a} \in \mu(q)\} = \emptyset$, or
(2) $\{\mu \in \mathit{sub}(q,D)~:~\atom{a} \in \mu(q)\} =
         \{\mu \in \mathit{sus}(q,D)~:~\atom{a} \in \mu(q)\}$.
\end{proposition}
\spaceAfterEnv
%
By combining Theorem~\ref{thm:suffUnfoundedSub}
and Proposition~\ref{prop:stronglyRedundant},
we have a declarative (yet inductive) specification of all the
strongly redundant components of $D$.
Importantly, the process of
identifying strongly redundant components and strongly unfounded substitutions
by exhaustively applying Theorem~\ref{thm:suffUnfoundedSub}
and Proposition~\ref{prop:stronglyRedundant}
is monotone and reaches a fixed-point (after no more than $|\mathit{comp}(D,\Sch)|$ steps) when no more key component can be
marked as strongly redundant.%

\spaceBeforeSubSec
\subsection{Idle Attributes}
\spaceAfterSubSec

Previously, we have described a technique to reduce inconsistencies
by progressively eliminating key components that are involved in
query substitutions but are redundant.
In the following, we show how to reduce inconsistencies
by reducing the cardinality of conflicting components, which in some cases
can be even treated as safe components.

The act of \emph{removing} an attribute $r[i]$ from
a triple $\langle q, D, \Sch\rangle$ consists of
reducing the arity of $r$ by one,
cutting down the $i$-th term of each $r$-atom of $D$ and $q$, and
adapting the positions of the primary key of $r$ accordingly.
Moreover, let $\mathit{attrs}(\Sch) = \{r[i]~|~r \in \R~\textrm{ and }~i \in [\alpha(r)]\}$,
let $B \subseteq \mathit{attrs}(\Sch)$, and let $A = \mathit{attrs}(\Sch) \setminus B$.
The \emph{projection} of $\langle q, D, \Sch\rangle$ on $A$, denoted by $\Pi_{A}(q, D, \Sch)$,
is the triple that is obtained from $\langle q, D, \Sch\rangle$ by removing all the attributes of $B$.
%
Consider a CQ $q$ and a predicate $r \in \R$.
The attribute $r[i]$ is {\em relevant} (w.r.t. $q$)
if $q$ contains an atom of the form $r(t_1,\ldots,t_{\alpha(r)})$
such that at least one of the following conditions is satisfied:
$(1)$ $i \in \kappa(r)$; or
$(2)$ $t_i$ is a constant; or
$(3)$ $t_i$ is a variable that occurs more than once in $q$; or
$(4)$ $t_i$ is a free variable of $q$.
An attribute which is not relevant is \emph{idle} (w.r.t. $q$).
An example is reported in \ref{sec:exampleIdle}.
%
%
The following theorem states that the consistent answer to a CQ
does not change after removing the idle attributes.

\spaceBeforeEnv
\begin{theorem}\label{thm:idle}
Consider a CQ $q$, the set
$R = \{r[i]\in \mathit{attrs}(\Sch)~|~r[i]~\textrm{ is relevant w.r.t. }q\}$,
and a database $D$.
It holds that $\mathit{ans}(q,D,\Sch) = \mathit{ans}(\Pi_{R}(q, D, \Sch))$.
\end{theorem}
\spaceAfterEnv

\spaceBeforeSubSec
\subsection{Conjunctive Queries and Safe Answers}
\spaceAfterSubSec
Let $\Sigma$ be a relational schema,
$D$ be a database, and $q = \exists \insY \, \varphi(\insX,\insY)$ be a CQ,
where we assume that $\mathit{\Sigma}$ contains only relevant attributes w.r.t. $q$
(idle attributes, if any, have been already removed).
Since $\mathit{ans}(q,D,\Sch) \subseteq q(D)$,
for each candidate answer $\tup_c \in q(D)$,
one should evaluate whether the BCQ $\bar{q} = \varphi(\tup_c,\insY)$ is (or is not)
consistently true in $D$.
Before constructing the conflict-join hypergraph of
$D$ (w.r.t. $\bar{q}$ and $\Sch$), however, one could check whether there is
a substitution $\mu$ that maps $\bar{q}$ to $D$ with the following property:
for each $\atom{a} \in \mu(\bar{q})$, the singleton $\{\atom{a}\}$ is a
safe component of $D$. And, if so, it is possible to conclude immediately
that $\tup_c \in \mathit{ans}(q,D,\Sch)$.
Intuitively, whenever the above property is satisfied, we say that
$\tup_c$ is a \emph{safe answer} to $q$ because, for each $R \in \mathit{rep}(D,\Sch)$,
it is guaranteed that $\mu(\bar{q}) \subseteq R$. The next result follows.

\spaceBeforeEnv
\begin{theorem}
Consider a CQ $q = \exists \insY \, \varphi(\insX,\insY)$, and
a tuple $\tup_c$ of $q(D)$.
If there is a substitution $\mu$ s.t. each atom of $\mu(\varphi(\tup_c,\insY))$
forms a safe component of $D$, then $\tup_c \in \mathit{ans}(q,D,\Sch)$.
\end{theorem}
\spaceAfterEnv

\newcounter{instr}
\setcounter{instr}{1}
\newcommand{\numi}{\arabic{instr}\addtocounter{instr}{1}}
\newcommand{\numb}{\textbf{\arabic{instr}\addtocounter{instr}{1}}}

\newcommand{\comm}[1]{\textcolor[rgb]{0.35,0.35,0.35}{\texttt{#1}}}
\newcommand{\before}{\vspace{10mm}}
\newcommand{\se}{\ensuremath{\textrm{ :-- }}}
\newcommand{\style}[1]{\ensuremath{\mathit{#1}}}
\newcommand{\vars}[1]{\ensuremath{\mathsf{#1}}}
\newcommand{\fun}[1]{\ensuremath{\texttt{#1}}}
\newcommand{\no}{\textrm{not }}
\newcommand{\coun}{\textrm{count}}
\newcommand{\sub}{\style{sub}}
\newcommand{\involvedAtom}{\style{involvedAtom}}
\newcommand{\confComp}{\style{confComp}}
\newcommand{\safeAns}{\style{safeAns}}
\newcommand{\subEq}{\style{subEq}}
\newcommand{\compEk}{\style{compEk}}
\newcommand{\inSubEq}{\style{inSubEq}}
\newcommand{\inCompEk}{\style{inCompEk}}
\newcommand{\redComp}{\style{redComp}}
\newcommand{\unfSub}{\style{unfSub}}
\newcommand{\residualSub}{\style{residualSub}}
\newcommand{\shareSub}{\style{shareSub}}
\newcommand{\ancestorOf}{\style{ancestorOf}}
\newcommand{\child}{\style{child}}
\newcommand{\keyCompInFrag}{\style{keyCompInFrag}}
\newcommand{\subInFrag}{\style{subInFrag}}
\newcommand{\frag}{\style{frag}}
\newcommand{\activeFrag}{\style{activeFrag}}
\newcommand{\activeAtom}{\style{activeAtom}}
\newcommand{\ignoredSub}{\style{ignoredSub}}

\spaceBeforeSec
\section{The Encoding in ASP}\label{sec:encoding}
\spaceAfterSec
In this section, we propose an ASP-based encoding to CQA that implements
the techniques described in Section \ref{sec:pruning}, and that is able to
deal directly with CQs, instead of evaluating separately the associated BCQs.
Hereafter, we assume the reader is familiar with Answer Set Programming~\cite{GeLi91,BrET11}
and with the standard syntax of ASP competitions~\cite{CaIR14}.
A nice introduction to ASP can be found in \cite{Ba02}, and in the ASP Core 2.0 specification in \cite{aspcore2}.
Given a relational schema $\Sigma = \langle \R, \alpha, \kappa \rangle$,
a database $D$, and a CQ $q = \exists \insY \, \varphi(\insX,\insY)$,
we construct a program $P(q,\Sigma)$ s.t. a tuple ${\bf t} \in q(D)$ belongs to
$\mathit{ans}(q,D,\Sch)$ iff each answer set of $D \cup P(q,\Sigma)$ contains an atom of the form $q^*(c,{\bf t})$, for some constant $c$.
%
%
%
Importantly, a large part of $P(q,\Sigma)$ does not depend on $q$ or $\Sigma$.
To lighten the presentation, we provide a simplified version
of the encoding that has been used in our experiments.
In fact, for efficiency reasons, idle attributes should be ``ignored on-the-fly''
without materializing the projection of $\langle q,D,\Sch\rangle$ on the relevant attributes; but this makes the encoding a little more heavy.
Hence, we first provide a naive way to consider only the relevant attributes,
and them we will assume that $\mathit{\Sigma}$ contains no idle attribute.
Let $R$ collect all the attributes of $\Sigma$ that are relevant w.r.t. $q$.
For each $r \in \R$ that occurs in $q$, let $\insW$ be a sequence of $\alpha(r)$ different variables
and $S = \{i~|~r[i]\in R\}$, the terms of the $r$-atoms of $D$ that are
associated to idle attributes can be removed via the rule $r'(\insW|_S) \se r(\insW)$.
Hereafter, let us assume that $\mathit{\Sigma}$ contains no idle attribute,
and $\insZ = \insX \cup \insY$.
Program $P(q,\Sigma)$ is depicted in Figure~\ref{fig:encoding}.

\begin{figure}[t!]
\hrule \vspace{-2.0mm} \footnotesize
\begin{tabbing}
\ \= \ \ \ \ \= \ \ \ \ \= \\
\> \comm{\%} \> \comm{Computation of the safe answer.}\\
\> \numi \> $\sub(\insZ) \se \varphi(\insZ)$. \\
\> \numi \> $\involvedAtom(\fun{k}_r(\hat{\tup}), \fun{nk}_r(\check{\tup})) \se
            \sub(\insZ), r(\tup)$. \` $\forall r(\tup) \in q$\\
\> \numi \> $\confComp(\vars{K}) \se \involvedAtom(\vars{K},\vars{NK}_1), \
            \involvedAtom(\vars{K},\vars{NK}_2), \ \vars{NK}_1 \neq \vars{NK}_2$.\\
\> \numi \> $\safeAns(\insX) \se \sub(\insZ),
             \no \confComp(\fun{k}_{r_1}(\hat{\tup}_1)), \ \ldots, \
             \no \confComp(\fun{k}_{r_n}(\hat{\tup}_n))$.\\
[2mm]\> \comm{\%} \> \comm{Hypergraph Construction.}\\
\> \numi \> $\subEq(\fun{sID}(\insZ), \fun{ans}(\insX)) \se
             \sub(\insZ), \ \no \safeAns(\insX)$.\\
\> \numi \> $\compEk(\fun{k}_r(\hat{\tup}), \vars{Ans}) \se
             \subEq(\fun{sID}(\insZ), \vars{Ans})$. \` $\forall r(\tup) \in q$\\
\> \numi \> $\inSubEq(\fun{atom}_r(\tup), \fun{sID}(\insZ)) \se
             \subEq(\fun{sID}(\insZ), \_)$. \` $\forall r(\tup) \in q$\\
\> \numi \> $\inCompEk(\fun{atom}_r(\tup), \fun{k}_r(\hat{\tup})) \se
             \compEk(\fun{k}_r(\hat{\tup}), \_), \
             \involvedAtom(\fun{k}_r(\hat{\tup}), \fun{nk}_r(\check{\tup}))$. \` $\forall r(\tup) \in q$\\
[2mm]\> \comm{\%} \> \comm{Pruning.}\\
\> \numi \> $\redComp(\vars{K},\vars{Ans}) \se \compEk(\vars{K},\vars{Ans}), \
             \inCompEk(\vars{A},\vars{K}),$ \\
\>       \>  \> $\#\coun\{\vars{S}: \inSubEq(\vars{A},\vars{S}), \
             \subEq(\vars{S},\vars{Ans})\} = 0$.\\
\> \numi \> $\unfSub(\vars{S},\vars{Ans}) \se
             \subEq(\vars{S},\vars{Ans}), \ \inSubEq(\vars{A},\vars{S}), \
             \inCompEk(\vars{A},\vars{K}), \redComp(\vars{K},\vars{Ans})$.\\
\> \numi \> $\redComp(\vars{K},\vars{Ans}) \se
             \compEk(\vars{K},\vars{Ans}),\ \inCompEk(\vars{A},\vars{K})$,\\
\>       \> \> $\#\coun\{\vars{S}: \inSubEq(\vars{A},\vars{S}), \subEq(\vars{S},\vars{Ans})\} =
               \#\coun\{\vars{S}: \inSubEq(\vars{A},\vars{S}), \unfSub(\vars{S},\vars{Ans})\}$.\\
\> \numi \> $\residualSub(\vars{S},\vars{Ans}) \se
             \subEq(\vars{S},\vars{Ans}), \ \no \unfSub(\vars{S},\vars{Ans})$.\\
[2mm]\> \comm{\%} \> \comm{Fragments identification.}\\
\> \numi \> $\shareSub(\vars{K}_1,\vars{K}_2,vars{Ans}) \se
             \residualSub(\vars{S},\vars{Ans}), \
             \inSubEq(\vars{A}_1,\vars{S}), \
             \inSubEq(\vars{A}_2,\vars{S}),$\\
\>       \>  \> $\vars{A}_1 \neq \vars{A}_2, \
                 \inCompEk(\vars{A}_1,\vars{K}_1), \
                 \inCompEk(\vars{A}_2,\vars{K}_2),
                 \vars{K}_1 \neq \vars{K}_2$.\\
\> \numi \> $\ancestorOf(\vars{K}_1,\vars{K}_2,\vars{Ans}) \se
             \shareSub(\vars{K}_1,\vars{K}_2,\vars{Ans}), \ \vars{K}_1 < \vars{K}_2$.\\
\> \numi \> $\ancestorOf(\vars{K}_1,\vars{K}_3,\vars{Ans}) \se
             \ancestorOf(\vars{K}_1,\vars{K}_2,\vars{Ans}), \
             \shareSub(\vars{K}_2,\vars{K}_3,\vars{Ans}), \vars{K}_1 < \vars{K}_3$.\\
\> \numi \> $\child(\vars{K},\vars{Ans}) \se \ancestorOf(\_,\vars{K},\vars{Ans})$.\\
\> \numi \> $\keyCompInFrag(\vars{K}_1, \fun{fID}(\vars{K}_1,\vars{Ans})) \se
              \ancestorOf(\vars{K}_1,\_,\vars{Ans}), \ \no \child(\vars{K}_1,\vars{Ans})$.\\
\> \numi \> $\keyCompInFrag(\vars{K}_2, \fun{fID}(\vars{K}_1,\vars{Ans})) \se
              \ancestorOf(\vars{K}_1,\vars{K}_2,\vars{Ans}), \ \no \child(\vars{K}_1,\vars{Ans})$.\\
\> \numi \> $\subInFrag(\vars{S},\fun{fID}(\vars{KF},\vars{Ans})) \se
             \residualSub(\vars{S},\vars{Ans}), \ \inSubEq(\vars{A},\vars{S}), \
             \inCompEk(\vars{A},\vars{K}),$ \\
\>       \> \> $\keyCompInFrag(\vars{K},\fun{fID}(\vars{KF},\vars{Ans}))$.\\
\> \numi \> $\frag(\fun{fID}(\vars{K},\vars{Ans}),\vars{Ans}) \se
             \keyCompInFrag(\_,\fun{fID}(\vars{K},\vars{Ans}))$.\\
[2mm]\> \comm{\%} \> \comm{Repair construction.}\\
\> \numi \> $1 \leqslant \{\activeFrag(\vars{F}):\frag(\vars{F},\vars{Ans})\} \leqslant 1
             \se \frag(\_,\_)$.\\
\> \numi \> $1 \leqslant \{\activeAtom(\vars{A}):\inCompEk(\vars{A},\vars{K})\} \leqslant 1 \se
              \activeFrag(\vars{F}), \keyCompInFrag(\vars{K},\vars{F})$.\\
\> \numi \> $\ignoredSub(\vars{S}) \se
              \activeFrag(\vars{F}), \
              \subInFrag(\vars{S},\vars{F}), \
              \inSubEq(\vars{A},\vars{S}), \
              \no \activeAtom(\vars{A})$.\\
[2mm]\> \comm{\%} \> \comm{New query.}\\
\> \numi \> $q^*(\fun{s},\vars{X}_1,\ldots,\vars{X}_n) \se
             \safeAns(\vars{X}_1,\ldots,\vars{X}_n)$.\\
\> \numi \> $q^*(\vars{F},\vars{X}_1,\ldots,\vars{X}_n) \se
            \frag(\vars{F},\fun{ans}(\vars{X}_1,\ldots,\vars{X}_n)), \
            \no \activeFrag(\vars{F})$.\\
\> \numi \> $q^*(\vars{F},\vars{X}_1,\ldots,\vars{X}_n) \se
             \activeFrag(\vars{F}), \ \subInFrag(\vars{S},\vars{F}), \
             \no \ignoredSub(\vars{S}),$\\
\>       \>  \> $\frag(\vars{F},\fun{ans}(\vars{X}_1,\ldots,\vars{X}_n))$.\\
\end{tabbing} \normalsize
\vspace{-5.0mm} \hrule \vspace{0.5mm} \vspace{-1mm}\caption{The Encoding in ASP.}\label{fig:encoding}\vspace{-5.0mm}
\end{figure}

\myParagraph{Computation of the safe answer.}
Via rule $1$, we identify the set $\mathcal{M} = \{\mu|_{\insZ}~:~\mu$ is a substitution and $\mu(\varphi(\insX,\insY))\subseteq D\}$.
%
It is now possible (rule $2$) to identify the atoms of $D$
that are involved in some substitution.
Here, for each atom $r(\tup) \in q$, we recall that $\hat{\tup}$ is the
subsequence of $\tup$ containing the terms in the positions of the primary key of $r$,
and we assume that $\check{\tup}$ are the terms of $\tup$ in the remaining positions.
In particular, we use two function symbols, $\fun{k}_r$ and $\fun{nk}_r$,
to group the terms in the key of $r$ and the remaining ones, respectively.
%
%
It is now easy (rule $3$) to identify the conflicting
components involved in some substitution.
%
%
%
Let $\varphi(\insX,\insY) = r_1(\tup_1),\ldots,r_n(\tup_n)$.
We now compute (rule $4$) the safe answers.
%

\myParagraph{Hypergraph construction.}
For each candidate answer $\tup_c \in q(D)$ that has not been
already recognized as safe, we construct the hypergraph
$H_D(\tup_c) = \langle D, E \rangle$ associated to the
BCQ $\varphi(\tup_c,\insY)$, where $E = E_q \cup E_\kappa$, as usual.
%
Hypergraph $H_D(\tup_c)$ is identified by the
functional term $\fun{ans}(\tup_c)$,
the substitutions of $E_q$ (collected via rule $5$) are identified by the set
$\{\fun{sID}(\mu(\insZ))~|~\mu \in \mathcal{M}$ and
    $\mu(\insX) = \tup_c\}$ of functional terms,
while the key components of $E_\kappa$ (collected via rule $6$) are identified by the set
$\{\fun{k}_r(\mu(\hat{\tup}))~|~\mu \in \mathcal{M}$ and
    $\mu(\insX) = \tup_c$ and
    $r(\tup) \in q\}$ of functional terms.
%
%
To complete the construction of the various hypergraphs,
we need to specify (rules $7$ and $8$) which are the atoms in each hyperedge.
%
%

\myParagraph{Pruning.}
We are now ready to identify (rules $9-11$)
the strongly redundant components and the strongly unfounded substitutions
(as described in Section \ref{sec:pruning})
to implement our cascade pruning mechanism.
Hence, it is not difficult to collect (rule $12$)
the substitutions that are not unfounded, that we call \emph{residual}.
%
%
%
%

\myParagraph{Fragments identification.}
Key components involving at least a residual
substitution (i.e., not redundant ones),
can be aggregated in fragments (rules $13-20$) by using the notion of
bunch introduced in Section \ref{sec:fragments}.
In particular, any given fragment $F$ -- associated to a candidate answer
$\tup_c \in q(D)$, and collecting the key components $K_1,\ldots,K_m$ --
is identified by the functional term $\fun{fID}(K_i,\tup_c)$
where, for each $j \in \{1,\ldots,m\} \setminus \{i\}$, the functional term
associated to $K_i$ lexicographically precedes the functional term
associated to $K_j$.
%
%
%
%
%
%

\myParagraph{Repair construction.}
%
Rules $1-20$ can be evaluated in polynomial time and have only one answer set,
while the remaining part of the program cannot in general.
In particular, rules $21-23$ generate the search space.
Actually, each answer set $M$ of $P(q,\Sigma)$ is associated (rule $21$)
with only one fragment, say $F$, that we call \emph{active} in $M$.
Moreover, for each key component $K$ of $F$, answer set $M$
is also associated (rule $22$) with only one atom of $K$, that
we also call \emph{active} in $M$.
Consequently, each substitution which involves atoms of $F$ but
also at least one atom which is not active, must be ignored in $M$ (rule $23$).
%
%

\myParagraph{New query.}
Finally, we compute the atoms of the form $q^*(c,{\bf t})$ via rules $24-26$.
\begin{figure}[t!]
    \centering
    \subfigure[Cactus plot.]{\label{fig:comp:cac}
        \includegraphics[height=4.3cm]{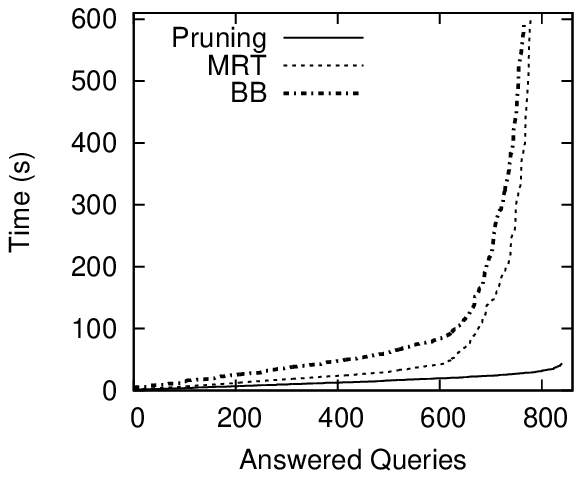}
    }
    \subfigure[Performance avg time and solved.]{\label{fig:comp:avgsol}
        \includegraphics[height=4.3cm]{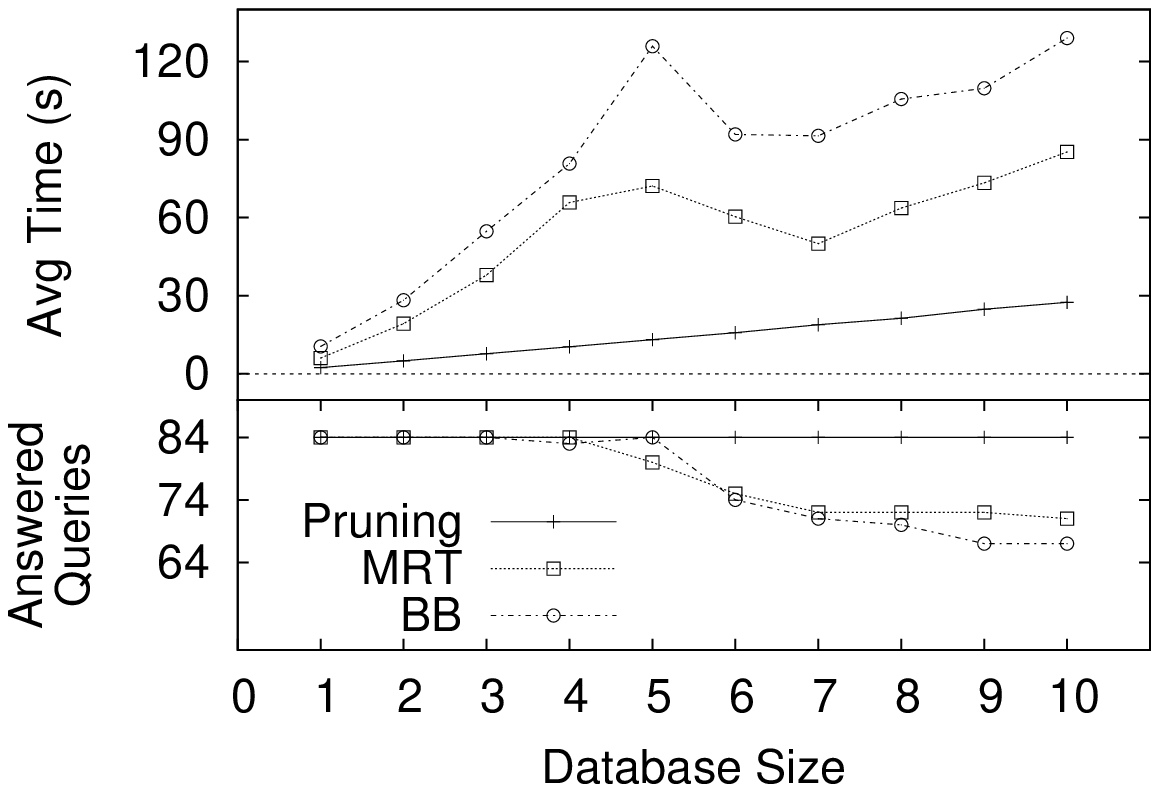}
    }
    \setcounter{subfigure}{0}\vspace{-0.2cm}
    \caption{Comparison with alternative encodings: answered queries and execution time.}\label{fig:comp}
    \vspace{-0.2cm}
\end{figure}


\spaceBeforeSec
\section{Experimental Evaluation}\label{sec:experiments}
\spaceAfterSec

The experiment for assessing the effectiveness of our approach is described in the following.
We first describe the benchmark setup and, then, we analyze the results.

\myParagraph{Benchmark Setup.}
The assessment of our approach was done using a benchmark employed in the literature for testing CQA systems on large inconsistent databases~\cite{KoPT13}.
%
It comprises 40 instances of a database schema with 10 tables, organized in four families of 10 instances each of which contains tables of size varying from 100k to 1M tuples; also it includes 21 queries of different structural features split into three groups depending on whether CQA complexity is coNP-complete (queries $Q_1, \cdots, Q_7$), PTIME but not FO-rewritable~\cite{Wijs09} (queries $Q_8, \cdots, Q_{14}$), and FO-rewritable (queries $Q_{15}, \cdots, Q_{21}$).
(See \ref{app:bench}).

We compare our approach, named  \pruning, with two alternative ASP-based approaches. 
In particular, we considered one of the first encoding of CQA in ASP that was introduced in~\cite{BaBe03}, and an optimized technique that was introduced more recently in \cite{MaRT13};
these are named \bertossi and \tplp, respectively. \bertossi and \tplp can handle a larger class of integrity constrains than \pruning, and only \tplp features specific optimization that apply also to primary key violations handling.
%
%
We constructed the three alternative encodings for all 21 queries of the benchmark, and we run them on the ASP solver \wasptwo~\cite{AlDR14nmr}, configured with the iterative coherence testing algorithm~\cite{AlDR14},
coupled with the grounder Gringo ver. 4.4.0 \cite{GeKKS11}.%
For completeness we have also run \clasp ver. 3.1.1~\cite{GeKS13} obtaining similar results. \wasp performed better in terms of number of solved instances on \tplp and \bertossi.
The experiment was run on a Debian server equipped with Xeon E5-4610 CPUs and 128GB of RAM. In each execution, resource usage was limited to 600 seconds and 16GB of RAM. Execution times include the entire computation, i.e., both grounding and solving.
All the material for reproducing the experiment (ASP programs, and solver binaries) can be downloaded from \url{www.mat.unical.it/ricca/downloads/mrtICLP2015.zip}.


\begin{figure}[t!]
    \centering  \def\figheight{3.9cm}
    \subfigure[Overhead (co-NP)]{\label{fig:ratio:ALLCONP}
        \includegraphics[height=\figheight]{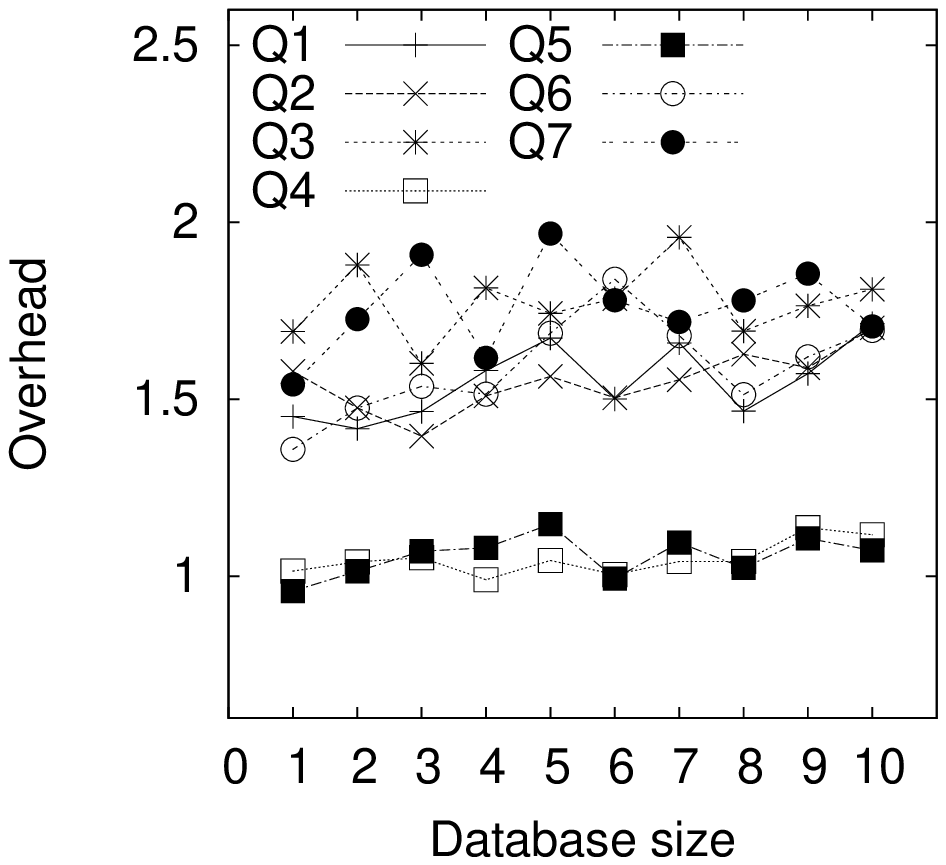}
    }
    \subfigure[Overhead (P)]{\label{fig:ratio:ALLP}
        \includegraphics[height=\figheight]{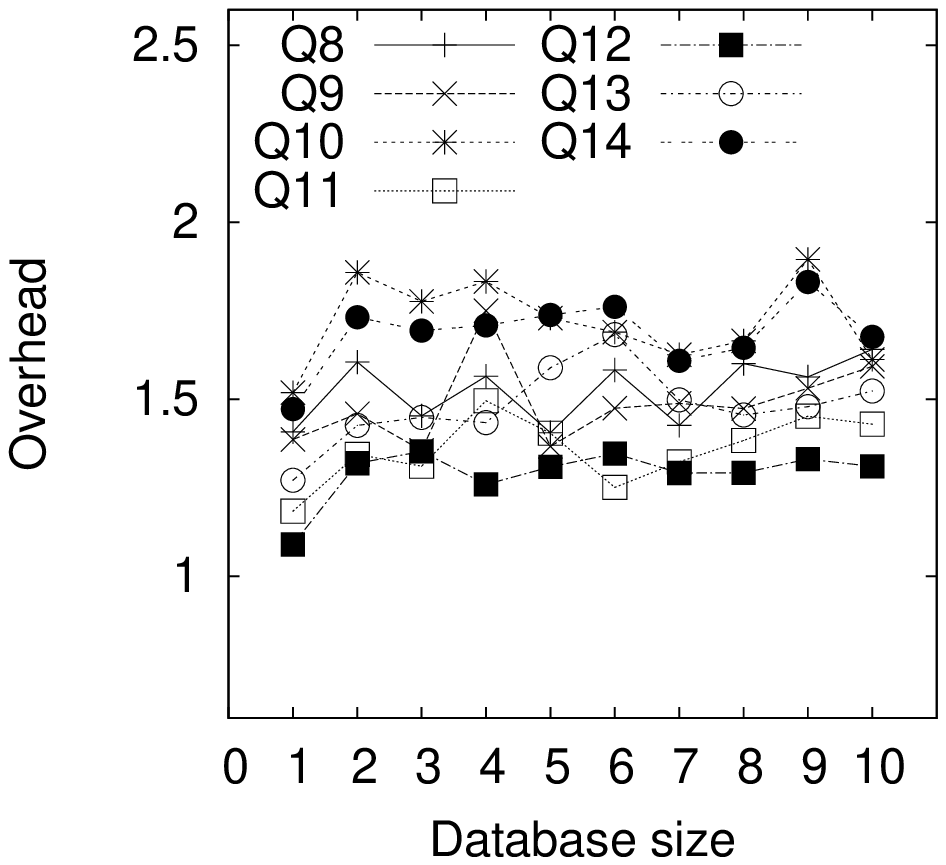}
    }
    \subfigure[Overhead (FO)]{\label{fig:ratio:ALLFO}
        \includegraphics[height=\figheight]{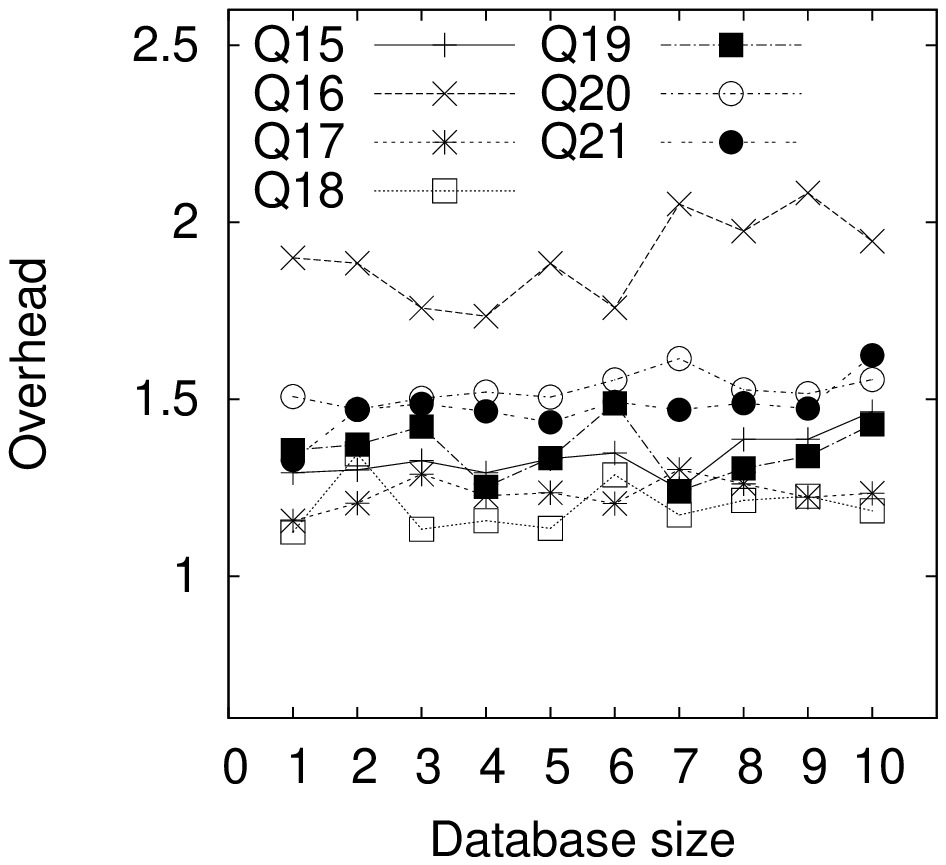}
    }\vspace{-0.2cm}
    \subfigure[Scalability (co-NP)]{\label{fig:scalability:CONP}
        \includegraphics[height=\figheight]{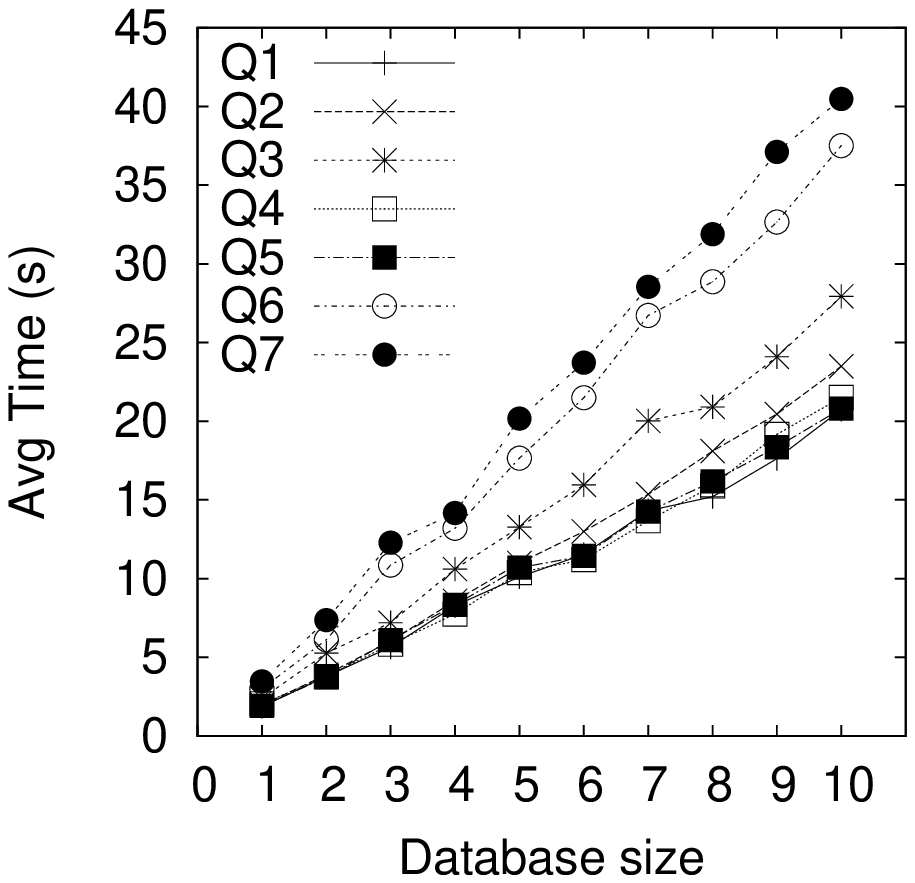}
    }
    \subfigure[Scalability (P)]{\label{fig:scalability:P}
        \includegraphics[height=\figheight]{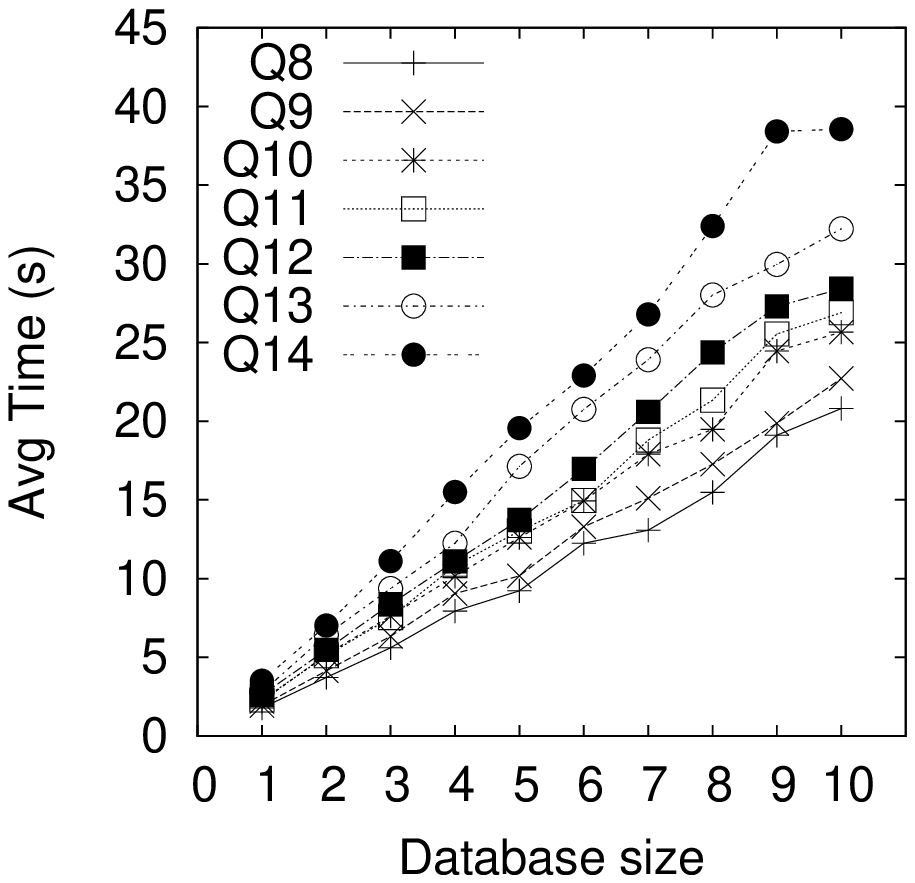}
    }
    \subfigure[Scalability (FO)]{\label{fig:scalability:FO}
        \includegraphics[height=\figheight]{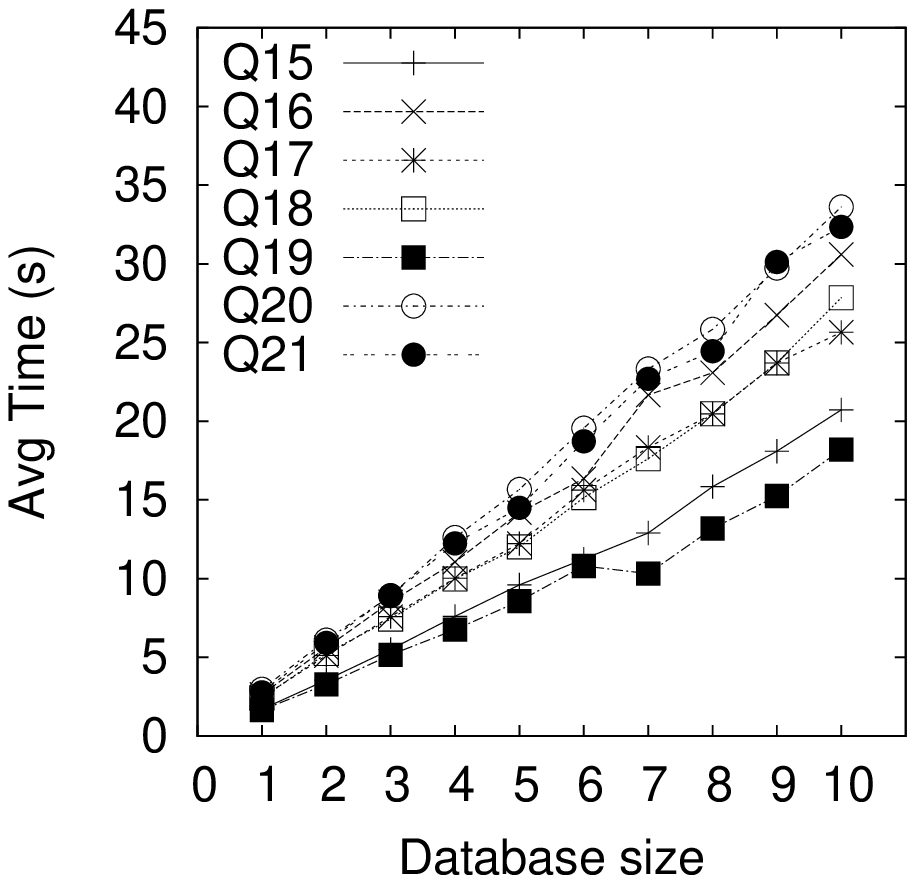}
    }
    \setcounter{subfigure}{0}
    \caption{Scalability and overhead of consistent query answering with Pruning encoding.}\label{fig:ratioscalability}
    \vspace{-0.4cm}
\end{figure}

\myParagraph{Analysis of the results.}
%
Concerning the capability of providing an answer to a query within the time limit,
we report that \pruning was able to answer the queries in all the 840 runs in the benchmark with an average time of 14.6s.
\tplp, and \bertossi solved only 778, and 768 instances within 600 seconds, with an average of 80.5s and 52.3s, respectively.
The cactus plot in Figure~\ref{fig:comp:cac} provides an aggregate view of the performance of the compared methods.
Recall that a cactus plot reports for each method the number of answered queries (solved instances) in a given time.
We observe that the line corresponding to \pruning in Figure~\ref{fig:comp:cac} is always below the ones of \tplp and \bertossi.
In more detail, \pruning execution times grow almost linearly with the number of answered queries, whereas \tplp and \bertossi show an exponential behavior. We also note that \tplp behaves better than \bertossi, and this is due to the optimizations done in \tplp that reduce the search space.

The performance of the approaches w.r.t. the size of the database is studied in Figure~\ref{fig:comp:avgsol}.
The x-axis reports the number of tuples per relation in tenth of thousands, in the upper plot is reported the number of queries answered in 600s, and in the lower plot is reported the corresponding the average running time.
We observe that all the approaches can answer all 84 queries (21 queries per 4 databases) up to the size of 300k tuples, then the number of answered queries by both \bertossi and \tplp starts decreasing. Indeed, they can answer respectively 74 and 75 queries of size 600k tuples, and only 67 and 71 queries on the largest databases (1M tuples). Instead, \pruning is able to solve all the queries in the data set.
The average time elapsed by running \pruning grows linearly from 2.4s up to 27.4s.
\tplp and \bertossi average times show a non-linear growth and peak at 128.9s and 85.2s, respectively.
(Average is computed on queries answered in 600s, this explains why it apparently
decreases when a method cannot answer some instance within 600s.)


The scalability of \pruning is studied in detail for each query in Figures~\ref{fig:ratioscalability}(d-f), 
each plotting the average execution times per group of queries of the same theoretical complexity.
It is worth noting that \pruning scales almost linearly in all queries, and independently from the complexity class of the query. This is because \pruning is able to identify and deal efficiently with the conflicting fragments.

We now analyze the performance of \pruning from the perspective of a measure called \textit{overhead}, which was employed in~\cite{KoPT13} for measuring the performance of CQA systems.
Given a query Q the overhead is given by $\frac{t_{cqa}}{t_{plain}}$, where $t_{cqa}$ is time needed for computing the consistent answer of Q, and $t_{plain}$ is the time needed for a plain execution of Q where the violation of integrity constraints are ignored.
Note that the overhead measure is independent of the hardware and the software employed, since it relates the computation of CQA to the execution of a plain query on the same system. Thus it allows for a direct comparison of \pruning with other methods having known overheads.
Following what was done in~\cite{KoPT13}, we computed the average overhead measured varying the database size for each query, and we report the results by grouping queries per complexity class in Figures~\ref{fig:ratioscalability}(a-c).
The overheads of \pruning is always below 2.1, and the majority of queries has overheads of around 1.5.
The behavior is basically ideal for query Q5 and Q4 (overhead is about 1).
The state of the art approach described in~\cite{KoPT13} has overheads that range between 5 and 2.8 on the very same dataset (more details on \ref{app:bench}). Thus, our approach allows to obtain a very effective implementation of CQA in ASP with an overhead that is often more than two times smaller than the one of state-of-the-art approaches.
We complemented this analysis by measuring also the overhead of \pruning w.r.t. the computation of safe answers, which provide an underestimate of consistent answers that can be computed efficiently (in polynomial time) by means of stratified ASP programs.
We report that the computation of the consistent answer with \pruning requires only at most 1.5 times more in average than computing the safe answer (detailed plots in \ref{app:bench}).
This further outlines that \pruning is able to maintain reasonable the impact of the hard-to-evaluate component of CQA.

\begin{figure}[t!]
    \centering
    \subfigure[Pruning]{\label{fig:hpr}
        \includegraphics[trim=2 0 0 0,clip,height=3.9cm]{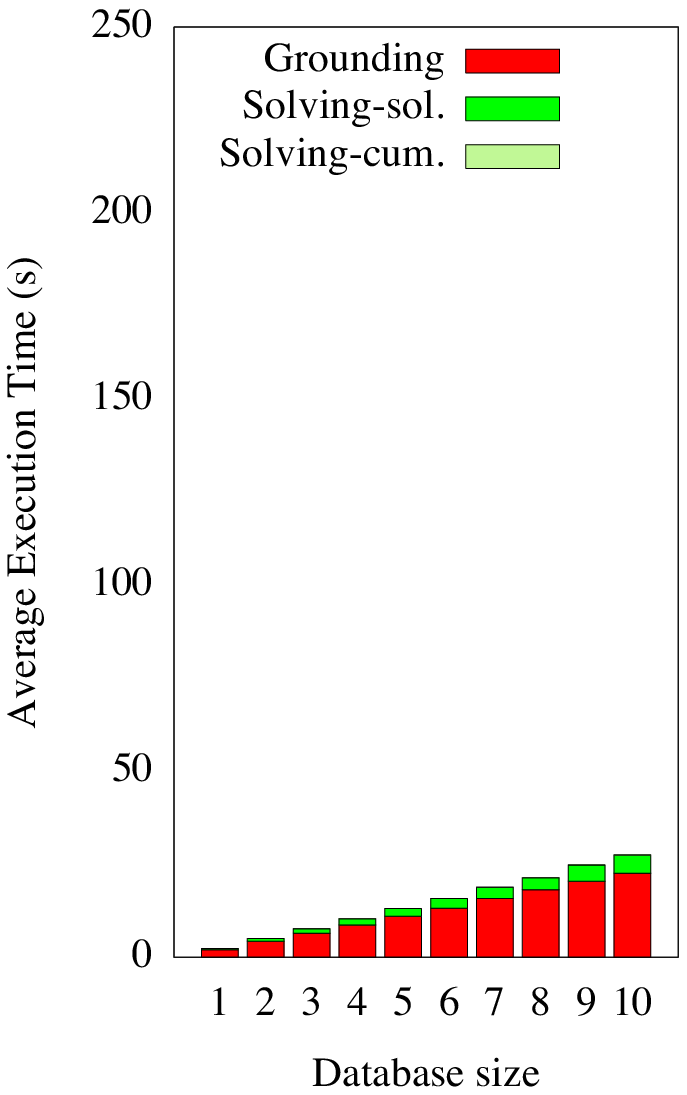}
    }
    \subfigure[BB]{\label{fig:hbb}
        \includegraphics[trim=2 0 0 0,clip,height=3.9cm]{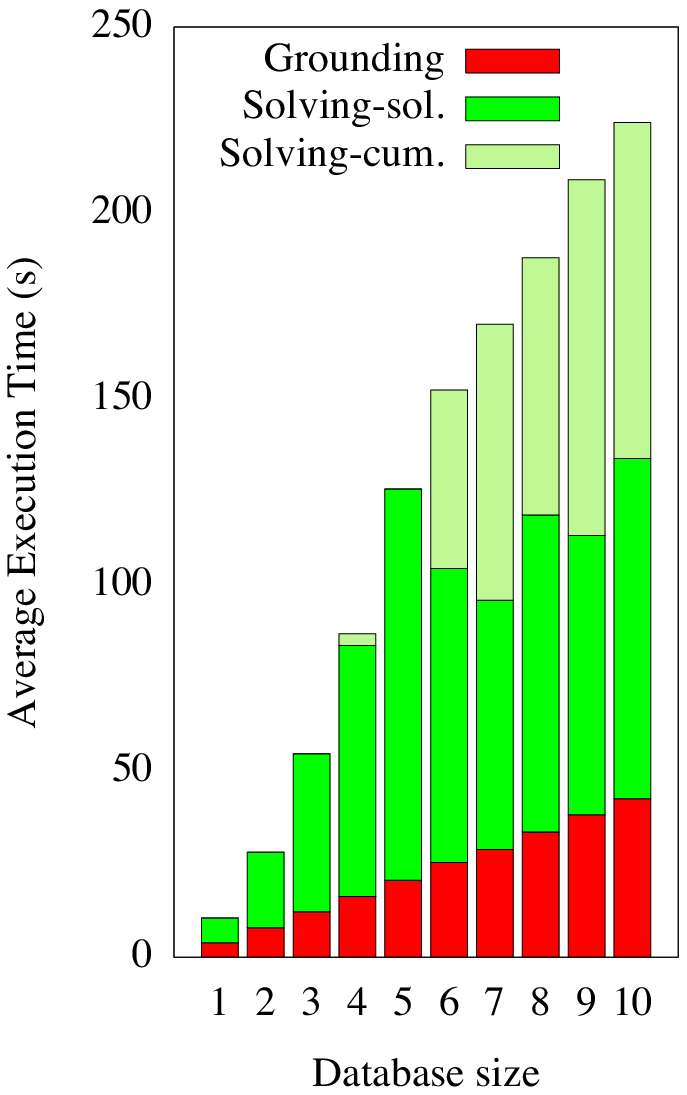}
    }
    \subfigure[MRT]{\label{fig:hmrt}
        \includegraphics[trim=2 0 0 0,clip,height=3.9cm]{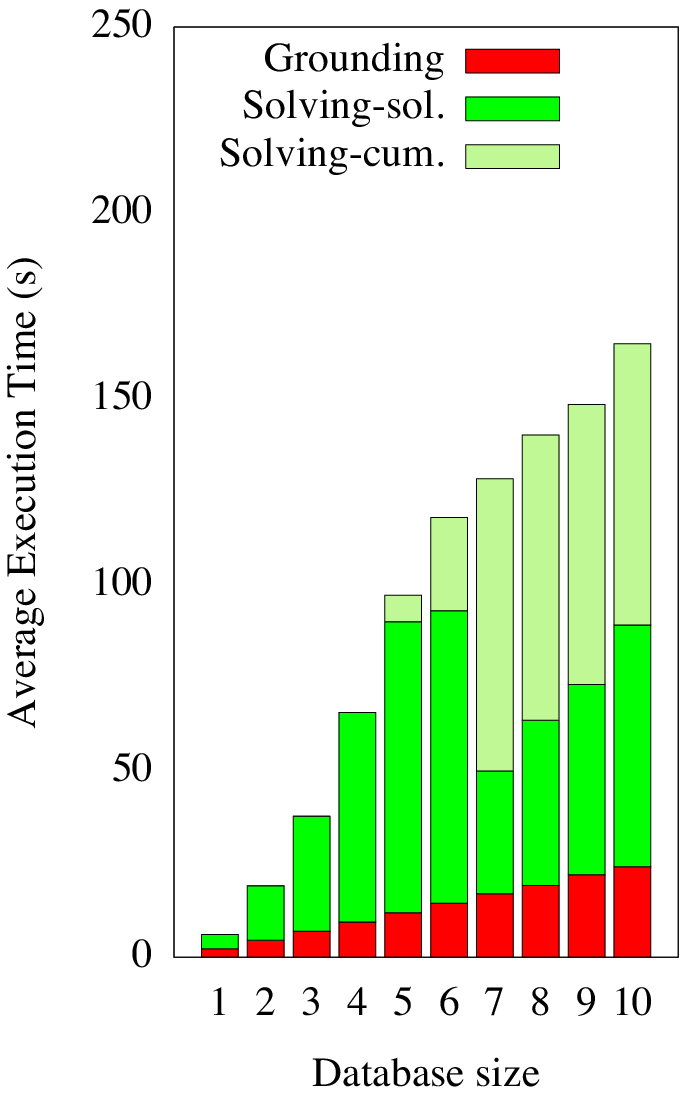}
    }
    \subfigure[Ground rules]{\label{fig:grperc}
        \includegraphics[trim=2 0 0 0,clip,height=3.85cm]{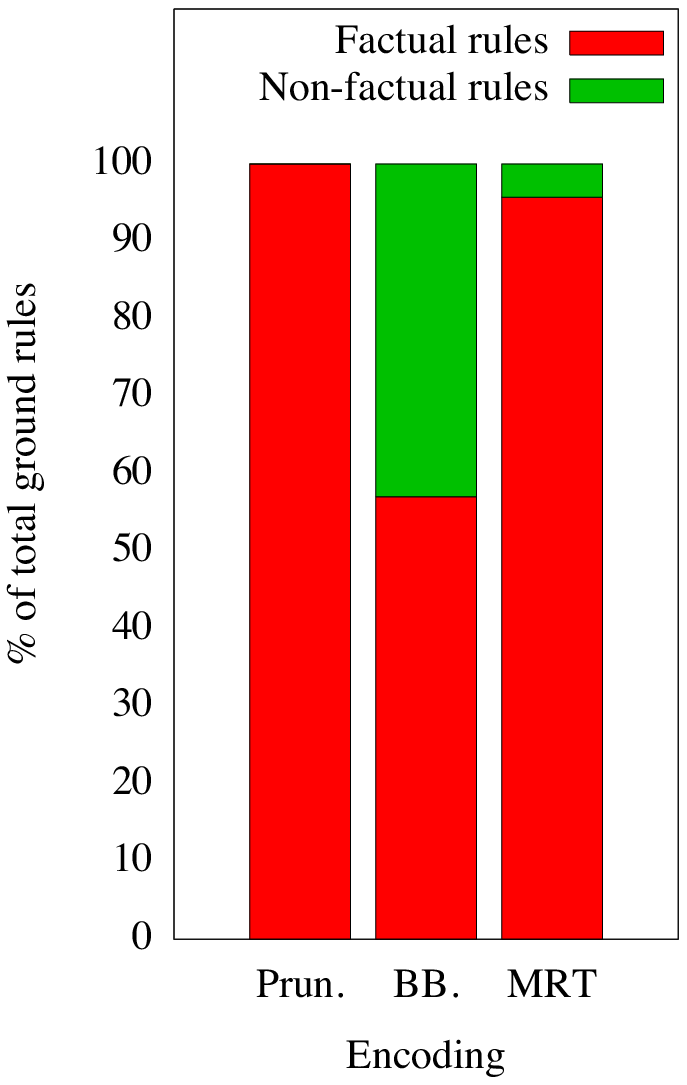}
    }
    \setcounter{subfigure}{0}\vspace{-0.2cm}
    \caption{Average execution times per evaluation step.}\label{fig:grounding}
    \vspace{-0.5cm}
\end{figure}

Finally, we have analyzed the impact of our technique in the various solving steps of the evaluation.
The first three histograms in Figure~\ref{fig:grounding} report the average running time spent for answering queries in databases of growing size for \pruning (Fig.~\ref{fig:hpr}), \bertossi (Fig.~\ref{fig:hbb}), and \tplp (Fig.~\ref{fig:hmrt}). In each bar different colors distinguish the average time spent for grounding and solving. In particular, the average solving time over queries \textit{answered within the timeout} is labeled Solving-sol, and each bar extends up to the average cumulative execution time computed over all instances, where each timed out execution counts 600s.
Recall that, roughly speaking, the grounder solves stratified normal programs, and the hard part of the computation is performed by the solver on the residual non-stratified program; thus, we additionally report in Figure~\ref{fig:grperc} the average number of facts (knowledge inferred by grounding) and of non-factual rules (to be evaluated by the solver) in percentage of the total for the three compared approaches.
The data in Figure~\ref{fig:grounding} confirm that most of the computation is done with \pruning during the grounding, whereas this is not the case for \tplp and \bertossi.
Figure~\ref{fig:grperc} shows that for \pruning the grounder produces a few non-factual rules (below 1\% in average),
whereas \tplp and \bertossi produce 5\%  and 63\% of non-factual rules, respectively.
Roughly, this corresponds to about 23K non-factual rules (resp., 375K non-factual rules) every 100K tuples per relation for \tplp (resp., \bertossi), whereas our approach produces no more than 650 non-factual rules every 100K tuples per relation.



\spaceBeforeSec
\section{Conclusion}
\spaceAfterSec \vspace*{-0.1cm}
Logic programming approaches to CQA were recently considered not competitive~\cite{KoPT13} on large databases affected by primary key violations.
In this paper, we proposed a new strategy based on a cascade pruning mechanism that dramatically reduces the number of primary key violations to be handled to answer the query. The strategy is encoded naturally in ASP, and
an experiment on benchmarks already employed in the literature demonstrates that our ASP-based approach is efficient on large datasets, and performs better than state-of-the-art methods in terms of overhead.
As far as future work is concerned, we plan to extend the \pruning method for handling inclusion dependencies, and other tractable classes of tuple-generating dependencies.

\newpage

\appendix\label{sec:appendix}

\section{Proofs}\label{sec:app-proofs}

Here we report the proofs of Theorems and Propositions reported in Section~\ref{sec:pruning}.

\subsection{- Proof of Proposition \ref{prop:fragment}}

Let us assume that $F_1 \models_\Sch q$.
This means that $q$ is true in every repair of $F_1$.
Since, by definition, for each repair $R_2$ of $F_2$, there exists a repair
$R_1$ of $F_1$ such that $R_1 \subseteq R_2$,
we conclude that $q$ must be true also in every repair of $F_2$.

\subsection{- Proof of Theorem \ref{thm:bunches}}

We we will prove the contrapositive.
To this end, let $B_1,\ldots,B_k$ be the bunches of $H_D$.
Assume that, for each $i\in[k]$, $B_i \not\models_\Sch q$.
This means that, for each $i \in [k]$, there exists a repair $R_i \in \mathit{rep}(B_i,\Sch)$
such that $R_i \not\models q$.
Consider now the instance $R = \bigcup_{i \in [k]} R_i$.
Since $B_1,\ldots,B_k$ always form a partition of $D$,
since for each $\mu \in \mathit{sub}(q,D)$, $\mu(q)$ is entirely contained in exactly one bunch, and
since each key component of $D$ is entirely contained in exactly one bunch,
we conclude that $R$ is a repair of $D$ and $R \not\models q$.
Hence $D \not\models_{\Sch} q$.

\subsection{- Proof of Proposition \ref{prop:TrueQueryHasNonRedundantComponent}}

$(\Rightarrow)$ If $D \not\models_\Sch q$, then by Proposition \ref{prop:fragment}
we have that, for each fragment $F$ of $D$,
$F \not\models_\Sch q$.
Moreover, by rephrasing Definition~\ref{def:redundant}, we have that
any key component $K$ of $D$ is redundant if the following condition is satisfied:
for each fragment $F$ of $D$, $F \not\models_\Sch q \vee F\setminus K \models_\Sch q$.
Hence, by combining the two, we conclude that each key component of $D$ is redundant.

$(\Leftarrow)$ If each key component $K$ of $D$
is redundant, by Proposition~\ref{prop:redundancyClosure}, we can conclude that
$D \not\models_\Sch q$, since the empty database cannot entail $q$.

\subsection{- Proof of Theorem \ref{thm:pruning}}

Let $F$ be a fragment of $D$ such that $F \models_\Sch q$.
%
%
By considering $F$ as a database and by Theorem \ref{thm:bunches},
we have that there exists at least a bunch $B$ of
the conflict-join hypergraph $H_F$ of $F$
such that $B \models_\Sch q$.
If $K \cap B = \emptyset$, then $F\setminus K \supseteq B$, and therefore,
by Proposition \ref{prop:fragment},
since $B$ is a fragment of $F\setminus K$, we have that $F\setminus K \models_\Sch q$.
If $K \subseteq B$, then let us consider one of the atoms $\atom{a} \in K$
that is not involved in any substitution.
But since $q$ is true in every repair of $B$ containing $\atom{a}$,
this means that $q$ is true also in every repair of $B \setminus K$.
And since $B \setminus K$ is a fragment of $F \setminus K$, also in this case
we can conclude that $F\setminus K \models_\Sch q$.

\subsection{- Proof of Theorem \ref{thm:suffUnfoundedSub}}

Let $K$ be a redundant component of $D$, and
$\mu$ be a substitution of $\mathit{sub}(q,D)$ such that $\mu(q) \cap K \neq \emptyset$.
Moreover, let $F$ be a fragment of $D$ such that $F \models_\Sch q$.
Since $K$ is redundant, by Definition~\ref{def:redundant},
we have that $F\setminus K \models_\Sch q$.
But since $\mu(q)$ necessarily contains an atom of $K$,
this means that for each repair $R \in \mathit{rep}(F\setminus K,\Sch)$,
there exists a substitution $\mu' \in \mathit{sub}(q,R)$ different from $\mu$
such that $\mu'(q) \subseteq R$.
But since the union of all these substitutions different from $\mu$
can be also used to entail $q$ in every repair of $F$,
by Definition~\ref{def:unfoundedSub},
we can conclude that $\mu$ is unfounded.

\section{- Example of relevant and idle attributes}\label{sec:exampleIdle}

Consider, for example, the schema $\Sch = \langle \R, \alpha, \kappa \rangle$,
where $\R = \{r_1,r_2\}$, $\alpha(r_1) = 3$, $\alpha(r_2) = 2$, and
$\kappa(r_1) = \kappa(r_2) = \{1\}$.
Consider also the database $D = \{r_1(1,2,3),$ $r_1(1,2,4),$ $r_2(2,5)\}$,
and the BCQ $q \equiv \exists X \exists Y \, r_1(1,X,Y), r_2(X,5)$.
The key components of $D$ are
$K_1 = \{r_1(1,2,3),$ $r_1(1,2,4)\}$ and $K_2 = \{r_2(2,5)\}$,
%
%
while the repairs of $D$ and $\Sch$ are
$R_1 = \{r_1(1,2,3),$ $r_2(2,5)\}$ and
$R_2 = \{r_1(1,2,4),$ $r_2(2,5)\}$.
Moreover, the set $\mathit{sub}(q,D)$ contains substitutions
$\mu_1 = \{X \mapsto 2, Y \mapsto 3\}$ and
$\mu_2 = \{X \mapsto 2, Y \mapsto 4\}$.
Finally, since $\mu_1$ maps $q$ to $R_1$,
and $\mu_2$ maps $q$ to $R_2$,
we can conclude that $D \models_\Sch q$.
However, one can observe that $K_1$ could be considered as a safe component with
respect to $q$.
In fact, variable $Y$ of $q$ -- being in a position that does not belong to $\kappa(r_1)$ --
occurs only once in $q$. And this intuitively means that whenever there exists a
substitution that maps $q$ in a repair containing $r_1(1,2,3)$, there must exist also a
substitution that maps $q$ in a repair containing $r_1(1,2,4)$.
Therefore, to avoid that $K_1$ produces two repairs,
one can consider only the first two attributes of $r_1$
and modify $q$ accordingly.
Hence, we can consider $\Sch' = \langle \R', \alpha', \kappa' \rangle$,
where $\R' = \{r_1',r_2\}$, $\alpha'(r_1') = \alpha'(r_2) = 2$ and $\kappa'(r_1') = \kappa'(r_2) = \{1\}$,
the database $D' = \{r_1'(1,2),$ $r_2(2,5)\}$, and
the BCQ $q' \equiv \exists X \exists Y \, r_1'(1,X), r_2(X,5)$.
Clearly, $D'$ is now consistent and entails $q'$.



\section{- Details on Benchmarks }\label{app:bench}
The benchmark considered in the paper was firstly used in \cite{KoPT13}.
It comprises several instances of varying size of a synthetic database specifically conceived to simulate reasonably high selectivities of the joins and a large number of potential answers.
Moreover it includes a set of queries of varying complexity and 40 instances of a randomly generated database. In the following we report the main characteristics of the data set and a link to an archive where the encoding and the binaries of the ASP system employed in the experiment can be also obtained.

\subsection{Queries}
It contains the following queries organized in groups depending on the respective complexity of CQA
(existential quantifiers are omitted for simplicity):

\begin{itemize}
\item \textbf{co-NP, not first-order rewritable}

\item[]$Q_{1}() = r_5(X, Y, Z), r_6(X_1, Y, W)$
\item[]$Q_{2}(Z) = r_5(X, Y, Z), r_6(X_1, Y, W)$
\item[]$Q_{3}(Z, W) = r_5(X, Y, Z), r_6(X_1, Y, W)$
\item[]$Q_{4}() = r_5(X, Y, Z), r_6(X_1, Y, Y), r_7(Y, U, D)$
\item[]$Q_{5}(Z) = r_5(X, Y, Z), r_6(X_1, Y, Y), r_7(Y, U, D)$
\item[]$Q_{6}(Z, W) = r_5(X, Y, Z), r_6(X_1, Y, W), r_7(Y, U, D)$
\item[]$Q_{7}(Z, W, D) = r_5(X, Y, Z), r_6(X_1, Y, W), r_7(Y, U, D)$

\item \textbf{PTIME, not first order rewritable}

\item[]$Q_{8}() = r_3(X, Y, Z), r_4(Y, X, W)$
\item[]$Q_{9}(Z) = r_3(X, Y, Z), r_4(Y, X, W)$
\item[]$Q_{10}(Z, W) = r_3(X, Y, Z), r_4(Y, X, W)$
\item[]$Q_{11}() = r_3(X, Y, Z), r_4(Y, X, W), r_7(Y, U, D) $
\item[]$Q_{12}(Z) = r_3(X, Y, Z), r_4(Y, X, W), r_7(Y, U, D) $
\item[]$Q_{13}(Z, W) = r_3(X, Y, Z), r_4(Y, X, W), r_7(Y, U, D)$
\item[]$Q_{14}(Z, W, D) = r_3(X, Y, Z), r_4(Y, X, W), r_7(Y, U, D)$

\item \textbf{First order rewritable}

\item[]$Q_{15}(Z) = r_1(X, Y, Z), r_2(Y, V, W)$
\item[]$Q_{16}(Z, W) = r_1(X, Y, Z), r_2(Y, V, W)$
\item[]$Q_{17}(Z) = r_1(X, Y, Z), r_2(Y, V), r_7(V, U, D)$
\item[]$Q_{18}(Z, W) = r_1(X, Y, Z), r_2(Y, V), r_7(V, U, D)$
\item[]$Q_{19}(Z) = r_1(X, Y, Z), r_8(Y, V, W)$
\item[]$Q_{20}(Z) = r_5(X, Y, Z), r_6(X_1, Y, W), r_9(X, Y, D)$
\item[]$Q_{21}(Z) = r_3(X, Y, Z), r_4(Y, X, W), r_10(X, Y, D)$

\end{itemize}

\subsection{Datasets}
We used exactly the same datasets employed in \cite{KoPT13}.
It comprises 40 samples of the same database, organized in four families of 10 instances each of which contains 10 tables of size varying from 100000 to 100000 tuples with increments 100000.
Quoting \cite{KoPT13}, the generation of databases has been done according with the following criterion:
''For every two atoms $R_i$, $R_j$ that share variables in any of the queries, approximately 25 of the facts in $R_i$ join with some fact in $R_j$ , and vice-versa. The third attribute in all of the ternary relations, which is sometimes projected out and never used as a join attribute in Table 1, takes values from a uniform distribution in the range $[1, r size/10]$. Hence, in each relation, there are approximately $r size/10$ distinct values in the third attribute, each value appearing approximately 10 times.''

\subsection{Encodings and Binaries}
We refrain from reporting here all the ASP encodings employed in the experiment since they are very lengthy.
Instead we report as an example the ASP program used for answering query Q7, and
provide all the material in an archive that can be downloaded from \url{www.mat.unical.it/ricca/downloads/mrtICLP2015.zip}.
The zip package also contains the binaries of the ASP system employed in the experiment.

\subsection{Pruning encoding of query Q7}
Let us classify the variables of $Q_7$:
\begin{itemize}
  \item All the variables are: $\{X,Y,Z,X_1,W,U,D\}$;
  \item The free variables are: $\{Z,W,D\}$;
  \item The variables involved in some join are: $\{Y\}$;
  \item The variables in primary-key positions are: $\{X,X_1,Y\}$;
  \item The variable in idle positions are: $\{U\}$
  \item The variable occurring in relevant positions are: $\{X,Y,Z,X_1,W,D\}$
\end{itemize}

\paragraph{Computation of the safe answer.}

\begin{alltt}\footnotesize
  sub(X,Y,Z,X1,W,D) \ :- \ r5(X,Y,Z), r6(X1,Y,W), r7(Y,U,D).
\end{alltt}\normalsize

\begin{alltt}\footnotesize
  involvedAtom(k-r5(X), nk-r5(V2,V3)) :- sub(X,Y,Z,X1,W,D), r5(X,V2,V3).
  involvedAtom(k-r6(X1), nk-r6(V2,V3)) :- sub(X,Y,Z,X1,W,D), r6(X1,V2,V3).
  involvedAtom(k-r7(Y), nk-r7(V3)) :- sub(X,Y,Z,X1,W,D), r7(Y,V2,V3).
\end{alltt}\normalsize

\begin{alltt}\footnotesize
  confComp(K) :- involvedAtom(K,NK1), involvedAtom(K,NK2), NK1 > NK2.
\end{alltt}\normalsize

\begin{alltt}\footnotesize
  safeAns(Z,W,D) :- sub(X,Y,Z,X1,W,D), not confComp(k-r5(X)),
                    not confComp(k-r6(X1)), not confComp(k-r7(Y)).
\end{alltt}\normalsize

\paragraph{Hypergraph construction.}

\begin{alltt}\footnotesize
  subEq(sID(X,Y,Z,X1,W,D), ans(Z,W,D)) :- sub(X,Y,Z,X1,W,D), not safeAns(Z,W,D).

  compEk(k-r5(X), Ans) :- subEq(sID(X,Y,Z,X1,W,D), Ans).
  compEk(k-r6(X1), Ans) :- subEq(sID(X,Y,Z,X1,W,D), Ans).
  compEk(k-r7(Y), Ans) :- subEq(sID(X,Y,Z,X1,W,D), Ans).
\end{alltt}\normalsize

\begin{alltt}\footnotesize
  inSubEq(atom-r5(X,Y,Z), sID(X,Y,Z,X1,W,D)) :- subEq(sID(X,Y,Z,X1,W,D), _).
  inSubEq(atom-r6(X1,Y,W), sID(X,Y,Z,X1,W,D)) :- subEq(sID(X,Y,Z,X1,W,D), _).
  inSubEq(atom-r7(Y,D), sID(X,Y,Z,X1,W,D)) :- subEq(sID(X,Y,Z,X1,W,D), _).

  inCompEk(atom-r5(X,V2,V3), k-r5(X)) :- compEk(k-r5(X), Ans),
                                         involvedAtom(k-r5(X), nk-r5(V2,V3)).
  inCompEk(atom-r6(X1,V2,V3), k-r6(X1)) :- compEk(k-r6(X1), Ans),
                                           involvedAtom(k-r6(X1), nk-r6(V2,V3)).
  inCompEk(atom-r7(Y,V3), k-r7(Y)) :- compEk(k-r7(Y), Ans),
                                      involvedAtom(k-r7(Y), nk-r7(V3)).
\end{alltt}\normalsize

\paragraph{Pruning.}

\begin{alltt}\footnotesize
  redComp(K,Ans) :- compEk(K,Ans), inCompEk(A,K),
                    #count\{S: inSubEq(A,S), subEq(S,Ans)\} = 0.

  unfSub(S,Ans) :- subEq(S,Ans), inSubEq(A,S), inCompEk(A,K), redComp(K,Ans).

  redComp(K,Ans) :- compEk(K,Ans), inCompEk(A,K),
                    X = #count\{S: inSubEq(A,S), subEq(S,Ans)\}
                    #count\{S: inSubEq(A,S), unfSub(S,Ans)\} >= X.

  residualSub(S,Ans) :- subEq(S,Ans), not unfSub(S,Ans).
\end{alltt}\normalsize

\paragraph{Fragments identification.}

\begin{alltt}\footnotesize
  shareSub(K1,K2,Ans) :- residualSub(S,Ans), inSubEq(A1,S), inSubEq(A2,S),
                         A1 <> A2, inCompEk(A1,K1), inCompEk(A2,K2), K1 <> K2.

  ancestorOf(K1,K2,Ans) :- shareSub(K1,K2,Ans), K1 < K2.
  ancestorOf(K1,K3,Ans) :- ancestorOf(K1,K2,Ans), shareSub(K2,K3,Ans), K1 < K3.

  child(K,Ans) :- ancestorOf(_,K,Ans).

  keyCompInFrag(K1, fID(K1,Ans)) :- ancestorOf(K1,_,Ans), not child(K1,Ans).
  keyCompInFrag(K2, fID(K1,Ans)) :- ancestorOf(K1,K2,Ans), not child(K1,Ans).

  subInFrag(S,fID(KF,Ans)) :- residualSub(S,Ans), inSubEq(A,S),
                              inCompEk(A,K), keyCompInFrag(K,fID(KF,Ans)).

  frag(fID(K,Ans),Ans) :- keyCompInFrag(_,fID(K,Ans)).
\end{alltt}\normalsize

\paragraph{Repairs Construction.}

\begin{alltt}\footnotesize
  1 <= \{activeFrag(F):frag(F,Ans)\} <= 1 :- frag(_,_).

  1 <= \{activeAtom(A):inCompEk(A,K)\} <= 1 :- activeFrag(F), keyCompInFrag(K,F).

  ignoredSub(S) :- activeFrag(F), subInFrag(S,F), inSubEq(A,S), not activeAtom(A).
\end{alltt}\normalsize

\paragraph{New query.}

\begin{alltt}\footnotesize
  q\(\sp{*}\)(s,Z,W,D) :- safeAns(Z,W,D).
  q\(\sp{*}\)(F,Z,W,D) :- frag(F,ans(Z,W,D)), not activeFrag(F).
  q\(\sp{*}\)(F,Z,W,D) :- activeFrag(F), subInFrag(S,F), not ignoredSub(S), frag(F,ans(Z,W,D)).
\end{alltt}\normalsize

\section{- Additional Plots } \label{app:exp}
We report in this appendix some additional plots.
In particular, we provide $(i)$ detailed plots for the overhead of \pruning w.r.t. safe answer computation;
$(ii)$ scatter plots comparing, execution by execution, \pruning with \bertossi and \tplp; and,
$(iii)$ an extract of \cite{KoPT13} concerning the overhead measured for the MIP-based approach for easing direct comparison with our results.

\begin{figure}[h!]\def\figheight{3.9cm}
    \subfigure[Pruning/Safe (co-NP)]{\label{fig:ratio:SAFECONP}
        \includegraphics[height=\figheight]{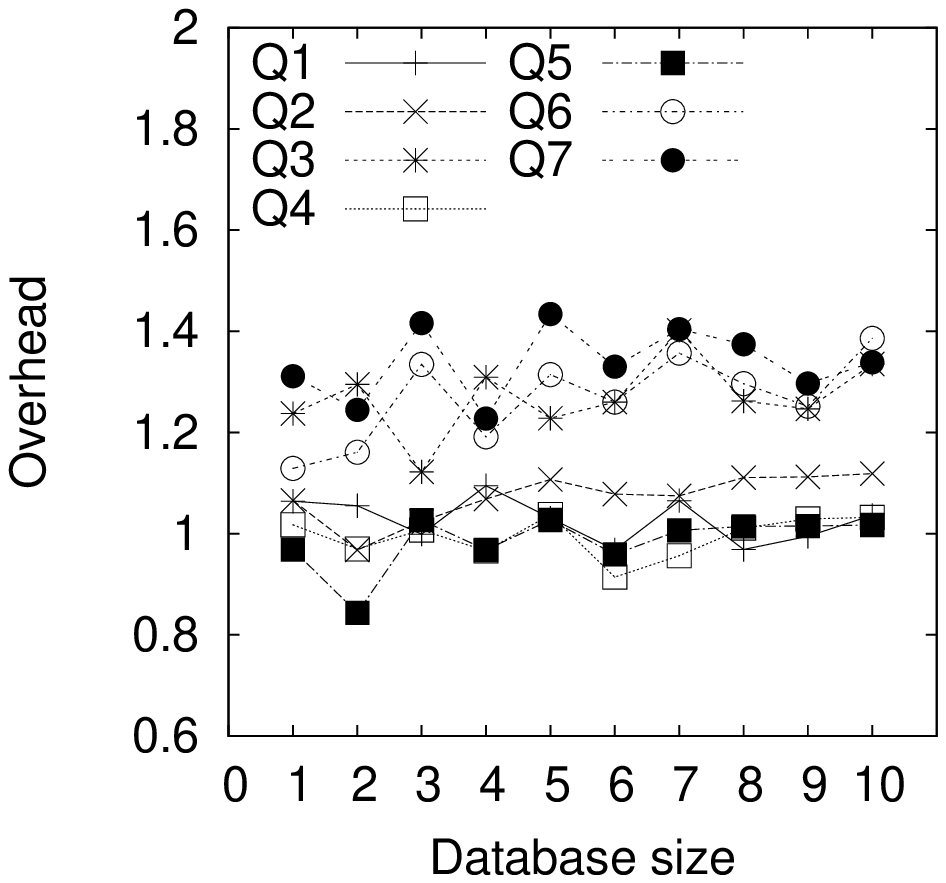}
    }
    \subfigure[Pruning/Safe (P)]{\label{fig:ratio:SAFEP}
        \includegraphics[height=\figheight]{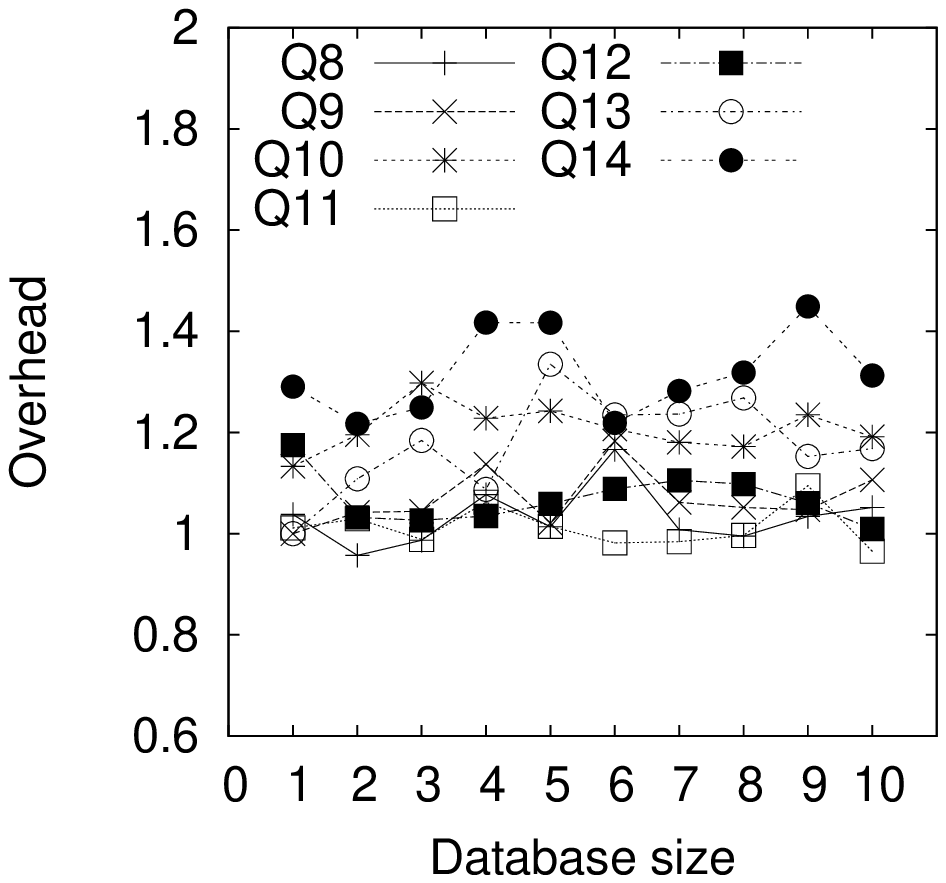}
    }
    \subfigure[Pruning/Safe (FO)]{\label{fig:ratio:SAFEFO}
        \includegraphics[height=\figheight]{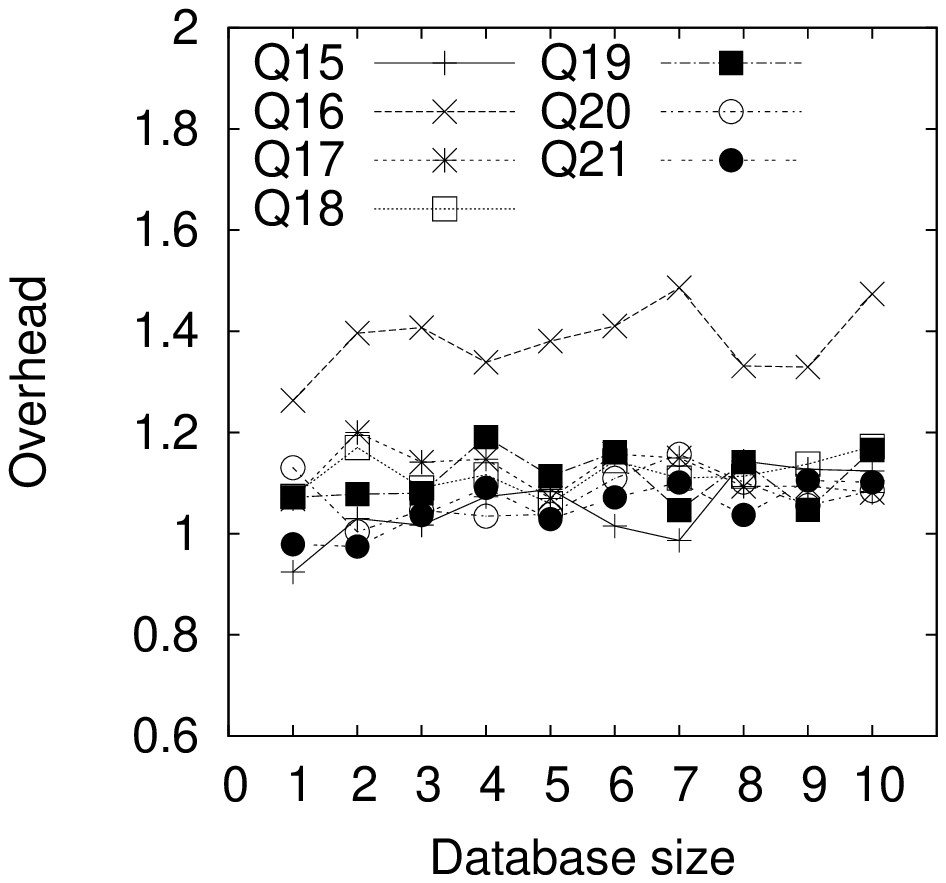}
    }
    \caption{Overhead of consistent query answering w.r.t. safe answers.}\label{fig:ratioSAFE}
    \vspace{-0.2cm}
\end{figure}

\paragraph{Overhead w.r.t. Safe Answers.}
We report in the following the detailed plots concerning the overhead of \pruning w.r.t. the computation of safe answers.
The results are reported in three plots grouping queries per complexity class in Figures~\ref{fig:ratioSAFE}.

It can be noted that the computation of consistent answers with \pruning takes at most to 1.5 times more than computing the safe answers in average, and is usually of about 1.2 times.

\begin{figure}[!h]
    \centering
    \subfigure[Pruning vs MRT.]{\label{fig:scatter:MRT}
        \includegraphics[height=5cm]{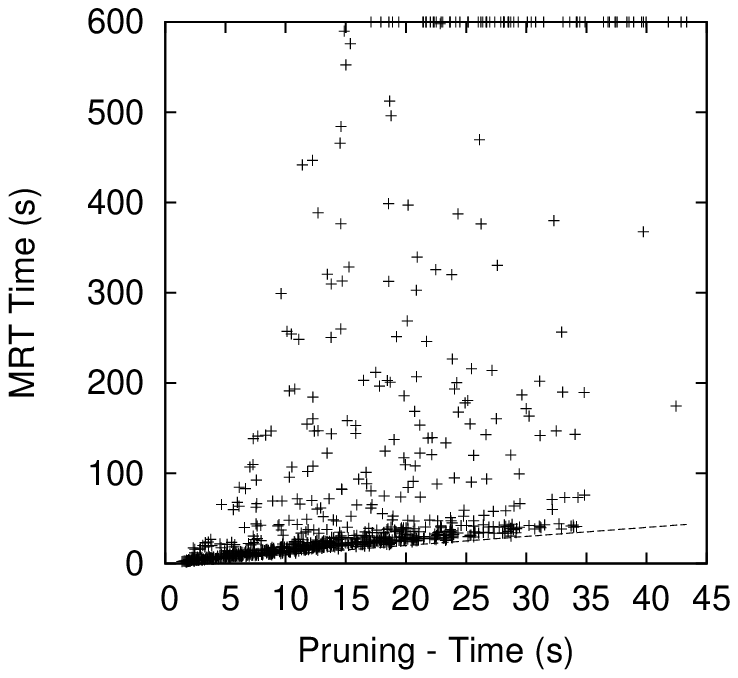}
    }
    \hspace{9mm}
    \subfigure[Pruning vs BB.]{\label{fig:scatter:ABC}
        \includegraphics[height=5cm]{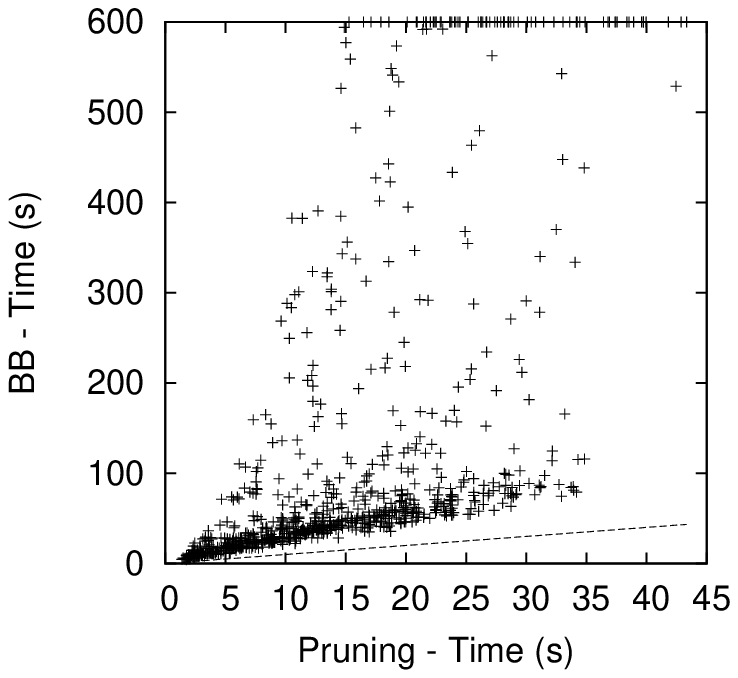}
    }
    \hspace{9mm}
    \subfigure[BB vs MRT.]{\label{fig:scatter:MRTABC}
        \includegraphics[height=5cm]{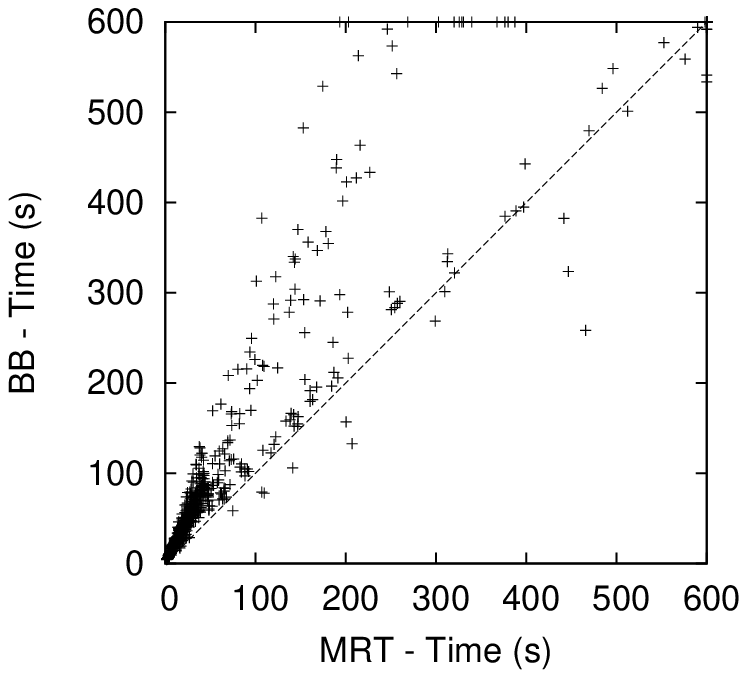}
    }
    \caption{Instance-wiese comparison with alternative encodings.}\label{fig:scatter}
\end{figure}

\paragraph{Scatter Plots.}
One might wonder what is the picture if the ASP-based approaches are compared instance-wise. An instance by instance comparison of \pruning with \bertossi and \tplp, is reported in the scatter plots in Figure~\ref{fig:scatter}.
In these plots a point $(x,y)$ is reported for each query, where $x$ is the running time of \pruning, and $y$ is the running time of \bertossi and \tplp, respectively in Figure~\ref{fig:scatter:ABC} and Figure~\ref{fig:scatter:MRT}.
The plots also report a dotted line representing the secant ($x=y$), points along this line indicates identical performance, points above the line represent the queries where the method on the $x$-axis performs better that the one in the $y$-axis and vice versa. Figure~\ref{fig:comp} clearly indicates that \pruning is also instance-wise superior to alternative methods.

\paragraph{Overhead of MIP approach from Kolaitis et. al (2013).}

\begin{figure}[h] \centering
\includegraphics[scale=0.7]{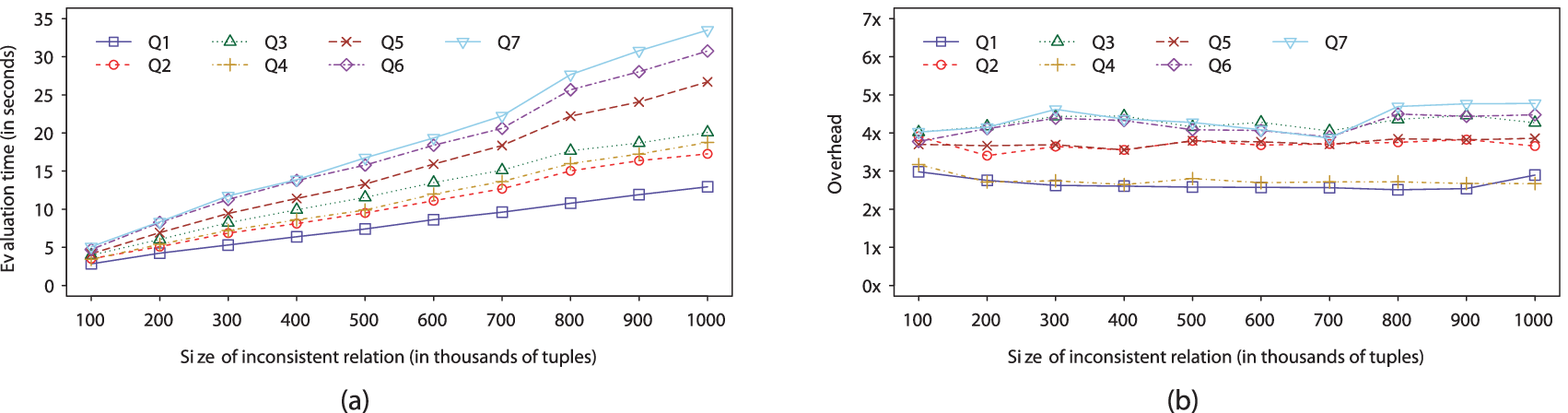}
\caption{Overhead of EQUIP for computing consistent answers of \textsf{coNP-hard} queries $Q_1$-$Q_7$.}
\label{fig:pema-coNP}
\end{figure}

\begin{figure}[h] \centering
\includegraphics[scale=0.7]{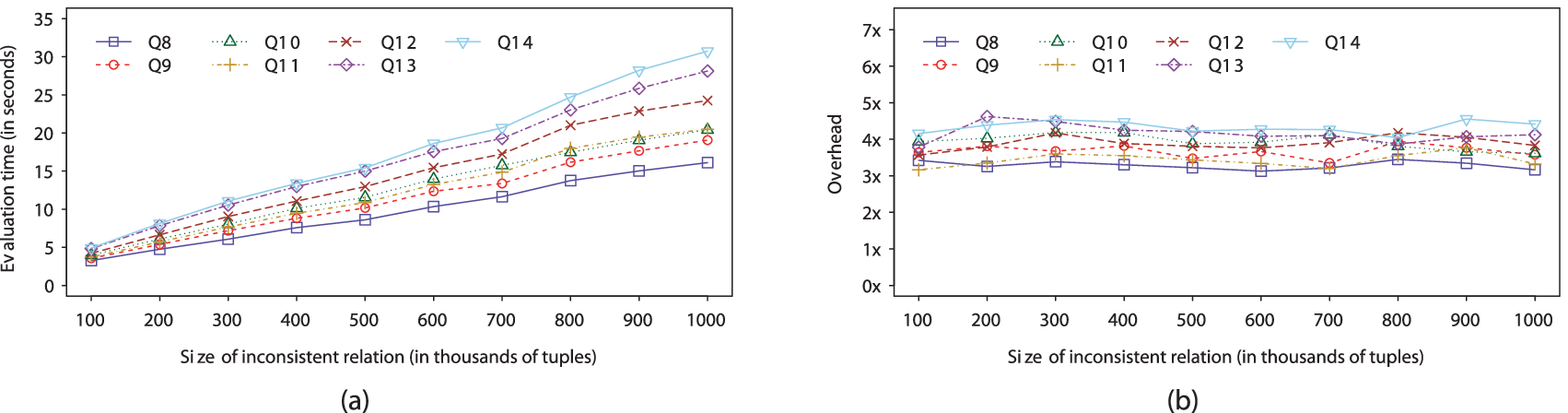}
\caption{Overhead of EQUIP for computing consistent answers of \textsf{PTIME}, but not-first-order rewritable queries $Q_8$-$Q_14$.}
\label{fig:pema-P}
\end{figure}

\begin{figure}[h] \centering
\includegraphics[scale=0.7]{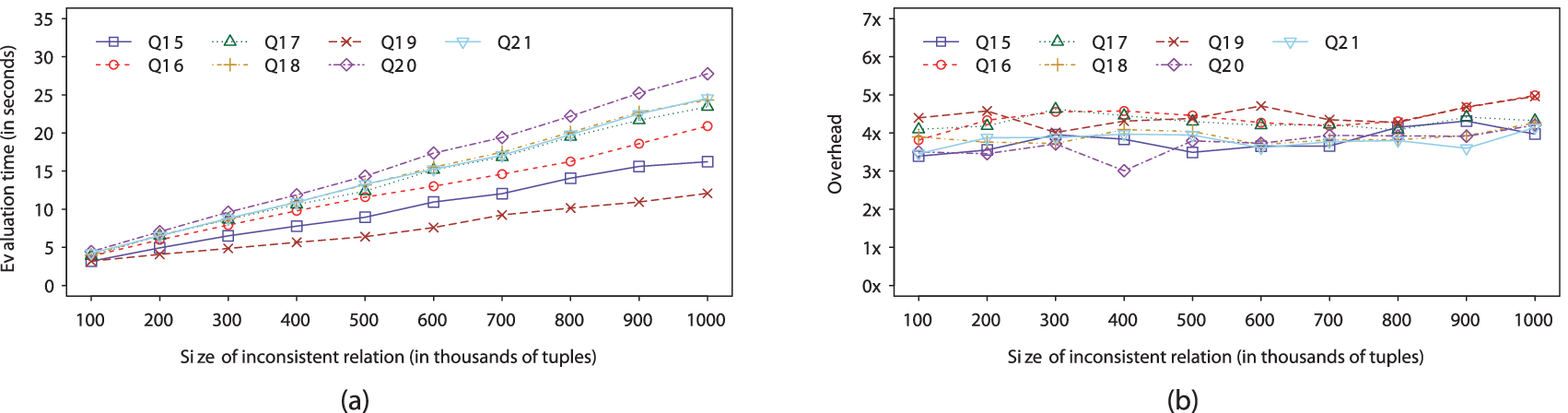}
\caption{Overhead of EQUIP for computing consistent answers of first-order rewritable queries $Q_{15}$-$Q_{21}$.}
\label{fig:pema-FO}
\end{figure}

\end{document}